\documentclass[lettersize]{IEEEtran}
\usepackage{amsmath,amsfonts}
\usepackage{algorithmic}
\usepackage{algorithm}
\usepackage{array}
\usepackage[caption=false,font=normalsize,labelfont=sf,textfont=sf]{subfig}
\usepackage{textcomp}
\usepackage{stfloats}
\usepackage{url}
\usepackage{verbatim}
\usepackage{subcaption}
\usepackage{graphicx}
\usepackage{cite}

\newcommand{\xmark}{$\times$}

\usepackage{multicol,booktabs,tabularx}
\newtheorem{theorem}{Theorem}
\newtheorem{definition}{Definition}[section]
\usepackage{bbold}
\usepackage{lineno}

\begin{document}

\title{Uncertainty-Guided Alignment for Unsupervised Domain Adaptation in Regression}

\author{Ismail Nejjar, Gaetan Frusque, Florent Forest, Olga Fink\\
Intelligent Maintenance and Operations Systems, EPFL, Lausanne, Switzerland}



\maketitle

\begin{abstract}
Unsupervised Domain Adaptation for Regression (UDAR) aims to adapt models from a labeled source domain to an unlabeled target domain for regression tasks. Traditional feature alignment methods, successful in classification, often prove ineffective for regression due to the correlated nature of regression features. To address this challenge, we propose Uncertainty-Guided Alignment (UGA), a novel method that integrates predictive uncertainty into the feature alignment process. UGA employs Evidential Deep Learning to predict both target values and their associated uncertainties. This uncertainty information guides the alignment process and fuses information within the embedding space, effectively mitigating issues such as feature collapse in out-of-distribution scenarios. We evaluate UGA on two computer vision benchmarks and a real-world battery state-of-charge prediction across different manufacturers and operating temperatures. Across 52 transfer tasks, UGA on average outperforms existing state-of-the-art methods. Our approach not only improves adaptation performance but also provides well-calibrated uncertainty estimates. The code is available in \url{https://github.com/ismailnejjar/UGA}.
\end{abstract}

\begin{IEEEkeywords}
Unsupervised Domain Adaptation, Regression, Uncertainty Estimation, Deep Learning, Computer Vision, Battery State of Charge
\end{IEEEkeywords}

\section{Introduction}\label{intro}

Effective transfer of regression models across diverse domains is crucial for applications such as computer vision and prognostics. However, this task remains challenging because variations in data distributions and feature relationships between domains can significantly degrade model performance when applied to new environments. In prognostics, domain gaps often arise from differences between industrial asset units,  operating conditions, and usage patterns, while in computer vision, these gaps are typically caused by variations in lighting, camera settings, and scene compositions.

Unsupervised Domain Adaptation (UDA) has emerged as a promising approach to address this challenge, allowing the transfer of knowledge from a labeled source domain to an unlabeled target domain~\cite{oza2023unsupervised,singhal2023domain}. While UDA has achieved significant success in classification tasks, its application to regression problems remains less explored and presents unique challenges~\cite{mansour2009domain}. In particular, traditional UDA methods designed for classification, such as correlation alignment~\cite{sun2016deep}, maximum mean discrepancy~\cite{borgwardt2006integrating,long2015learning}, and adversarial training~\cite{ganin2016domain}, have shown limited effectiveness when applied to regression tasks~\cite{chen2021representation}. Nevertheless, regression tasks are prevalent in various fields, including computer vision (e.g., depth estimation~\cite{piccinelli2023idisc,kong2024robodepth}, age prediction~\cite{wen2020adaptive,nejjar2024context}) and prognostics (e.g., remaining useful life prediction of complex industrial and infrastructure assets~\cite{fink2020potential,nejjar2024domain}).

The challenges of UDA in regression compared to classification arise from the fundamental differences between classification and regression tasks. 
The first challenge is related to the structure of the feature space. In classification,  learned features are typically distributed across various dimensions of the embedding space, forming distinct clusters for each class~\cite{hinton1984distributed,lecun2015deep}.  This rich and dispersed feature representation facilitates easier alignment between source and target domains, which is crucial for traditional alignment-based UDA methods. In contrast, regression tasks tend to generate more correlated features, with meaningful information concentrated in lower-dimensional subspaces. This phenomenon is often due to deep learning models developing  \emph{lazy} representations \cite{kunin2024get}, where the model relies on a limited subset of features to make predictions. These inherent differences require specialized approaches for UDAR, with recent state-of-the-art methods focusing on subspace alignment rather than aligning the entire feature space~\cite{chen2021representation,Nejjar_2023_CVPR}. By targeting these lower-dimensional informative subspaces, these methods better address the unique challenges posed by regression tasks.

A second major challenge in UDAR is linked to the quantification of prediction uncertainty. In classification tasks, models inherently can provide confidence estimates through class-wise probabilities~\cite{mukhoti2023deep}. Recent advancements in classification-based domain adaptation methods heavily rely on their confidence or entropy measures to generate accurate pseudo-labels for self-training~\cite{zou2018unsupervised,forest_calibrated_2023,hao_simplifying_2024}. In contrast, regression models typically output point estimates without associated uncertainty measures, presenting a unique challenge for UDAR.
\textit{Moreover, traditional regression loss functions, such as Mean Squared Error (MSE), inherently assume constant uncertainty across all data points. This assumption prevents the model from differentiating between in-distribution (confident) and out-of-distribution (unconfident) samples.}
Although there is no direct equivalent to classification confidence in regression, incorporating uncertainty estimates could serve a similar function. These uncertainty measures can help identify samples where the model is more confident, thereby guiding the alignment process during domain adaptation. In critical applications, understanding the confidence levels of predictions not only provides valuable insights for decision-making but could also be effective for domain adaptation strategies for regression tasks by focusing on reliable predictions. Leveraging uncertainty in regression thus holds the potential to improve the effectiveness of UDAR methods, ensuring more robust and trustworthy model performance across diverse domains.

Most well-known methods of uncertainty quantification in deep learning require multiple forward passes at test time, such as Monte Carlo Dropout (MCD)~\cite{gal2016dropout}. Among these, Deep Ensembles have generally performed best in uncertainty prediction~\cite{ovadia2019can}, but their significant memory and compute burden at training and test time hinders their adoption in real-life and mobile applications. Consequently, there has been increased interest in uncertainty quantification using deterministic single forward-pass neural networks, which have a smaller footprint and lower latency.
Two popular works in single forward-pass uncertainty, deterministic uncertainty quantification (DUQ)~\cite{van2020uncertainty} and Spectral-normalized Neural Gaussian Process (SNGP)~\cite{liu2020simple}, propose distance-aware output layers and introduce additional inductive biases in the feature extractor to encourage smoothness and sensitivity in the latent space. These methods perform well and are almost competitive with Deep Ensembles on Out-of-Distribution benchmarks. However, they require substantial changes to training and introduce additional hyper-parameters due to the specialized output layers used at training.
A simpler approach is Evidential Deep Learning~\cite{sensoy2018evidential,huang2023evidential,zhou2023trustworthy}, which aims to learn to predict the parameters of a higher-order distribution, allowing for direct uncertainty estimation without the need for multiple forward passes, and requires only a simple change in the loss function.

\begin{figure*}[htbp]
    \centering
    \begin{minipage}[t]{0.4\linewidth}
        \centering
        \includegraphics[width=\columnwidth]{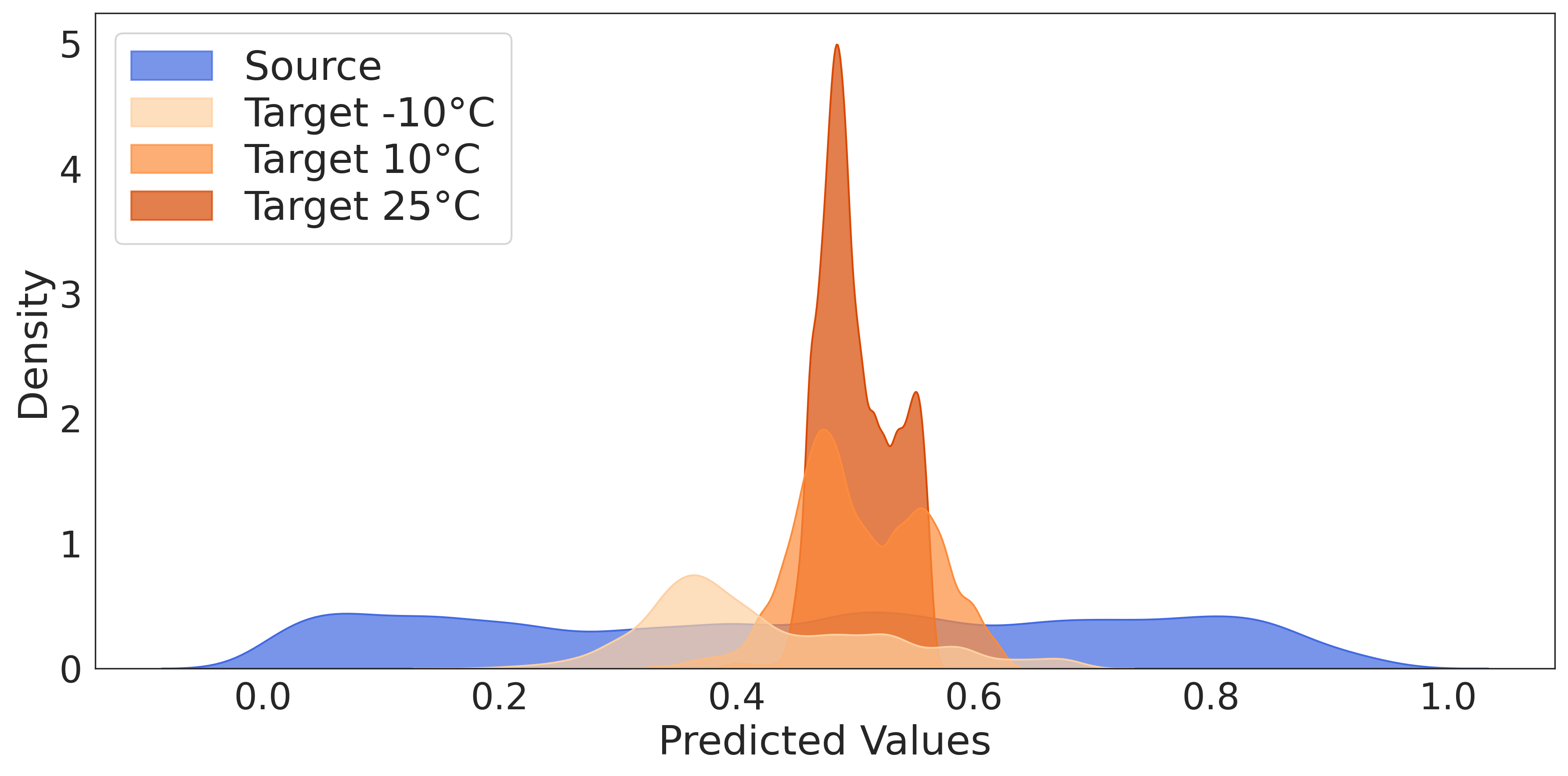}
        \refstepcounter{subfigure}
        \caption*{(\thesubfigure) Density plot comparing model predictions for battery State of charge between the source domain at -20°C and target domains of increasing temperature, showing the effect of domain shift on prediction patterns.}
        \label{fig:density_sub}
    \end{minipage}
    \hspace{20pt}
    \begin{minipage}[t]{0.4\linewidth}
        \centering
        \includegraphics[width=\columnwidth]{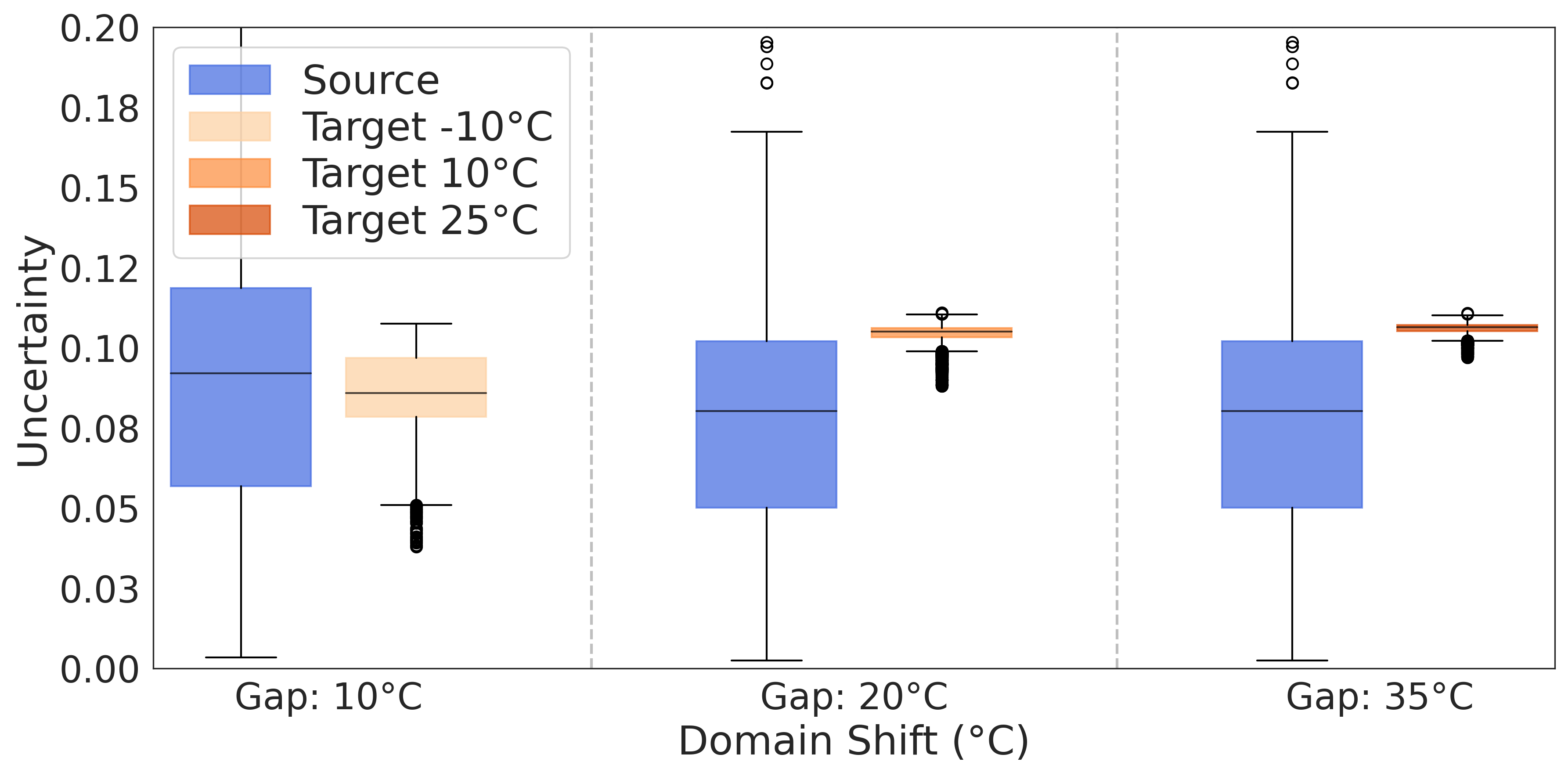}
        \refstepcounter{subfigure}
        \caption*{(\thesubfigure) Box plot comparing uncertainty distributions between source domain (-20°C) and target domains of increasing temperature, showing the impact of domain shift on model uncertainty.}
        \label{fig:box_plot_sub}
    \end{minipage}
    \caption{Visualization of domain shift effects on prediction distribution and uncertainty in regression tasks for battery state of charge prediction. (a) Density plot of model predictions. (b) Box plot of uncertainty distributions.}
    \label{fig:intro_fig}
    \vspace{-0.5cm}
\end{figure*}

To illustrate the challenges discussed above, we present two visualizations in Figure~\ref{fig:intro_fig}, where the domains are defined by battery operating temperature and the task is to predict the state of charge of a battery. Figure~\ref{fig:intro_fig} (a) illustrates the density of model predictions under varying degrees of domain shift. As the temperature difference (domain gap) increases,  the distribution of predicted values becomes narrower, indicating a tendency toward less diverse predictions for out-of-distribution data. This phenomenon, known as feature collapse~\cite{van2020uncertainty}, occurs when feature extractors map the entire target domain to a confined region within the source domain feature space. Figure~\ref{fig:intro_fig} (b) demonstrates the differences in uncertainty distributions between the source domain (-20°C) and target domains with progressively higher temperatures, using a model trained with Deep Evidential Regression~\cite{amini2020deep}. This visualization highlights the role of uncertainty as an indicator of domain shift in regression tasks. Notably, as the domain gap widens with the temperature increase, the uncertainty distributions become narrower, further suggesting feature collapse within the target domain.

The visualizations in Figure~\ref{fig:intro_fig} underscore these two key challenges in UDAR: (1) the need for effective feature alignment strategies that account for the unique characteristics of regression tasks, and (2) the importance of quantifying and leveraging uncertainty in the adaptation process. To address both challenges, we propose a novel approach named Uncertainty-Guided Alignment (UGA) for unsupervised domain adaptation in regression tasks. Our method leverages Evidential Deep Learning to effectively guide the adaptation process. In UGA, we use uncertainty information in two distinct ways:
\begin{enumerate}
    \item Guiding the alignment process: We focus on features that are more certain in the source domain and potentially transferable to the target domain.
    \item Embedding space regularization: We encourage the model to learn more robust and generalizable features.
\end{enumerate} 

We evaluate our method on standard computer vision regression tasks, including the dSprites and MPI3D datasets, demonstrating its effectiveness across different types of domain shifts. Additionally, we apply UGA to a real-world prognostics problem: battery state-of-charge estimation under varying temperature conditions and across different manufacturers. In this real-world scenario, we observe that the UGA model's performance remains consistently superior across different domain gaps compared to baseline methods. The model demonstrates well-calibrated uncertainty estimates throughout the battery state-of-charge, with higher uncertainty at the beginning of a new battery cycle that gradually decreases as the battery ages and predictions become more accurate. Across 52 transfer tasks, our proposed method outperforms existing state-of-the-art approaches on average, demonstrating its effectiveness in improving adaptation performance and providing valuable confidence measures for regression predictions in unsupervised domain adaptation tasks.

\section{Related Work}\label{related_work}

\subsection{Unsupervised Domain Adaptation} 
In the realm of UDA, two main directions have emerged. The first direction aims to reduce the discrepancy between domains by using optimal transport~\cite{courty_optimal_2017,fatras2021unbalanced,chang2022unified,ren2024towards} or matching moments of the distributions through the Maximum Mean Discrepancy (MMD)~\cite{long2015learning,chen2020homm, Luo_2021_CVPR,shi2023transfer}. The second direction in UDA revolves around learning domain-invariant representations through adversarial learning techniques~\cite{ganin2016domain,rangwani2022closer,chen2022reusing}. In the context of classification, leveraging self-training with confident pseudo-labels has proven to be effective~\cite{zou_domain_2018, sun2022safe, karim2023c, hoyer2023mic,huo2024domain,hao_simplifying_2024}. Furthermore, in the context of UDA, various mixing strategies have been developed to foster domain-invariant representations within the embedding space~\cite{xu2019adversarial, DBLP:journals/corr/abs-2007-03141,jing2023marginalized}, ultimately improving the model's generalization capabilities. A recent line of works leverage foundation models such as CLIP to perform the adaptation \cite{ge2023domain}.

The field of Unsupervised Domain Adaptation for Regression has received relatively less attention than classification. Early theoretical works have derived generalization bounds~\cite{mansour_domain_2009,cortes_domain_2011,redko_survey_2022}. 
More recently, \cite{wu2022distribution} proposed a framework for domain adaptation in regression, specifically designed for scenarios where target labels are accessible.  This framework encompasses popular approaches, including domain-invariant representation learning, reweighting, and adaptive Gaussian processes.
Recent endeavors have focused on addressing the unique challenges within the unsupervised setting for domain adaptation for regression~\cite{chen2021representation, Nejjar_2023_CVPR, nejjar2023domain,dhaini2023unsupervised,yang2025cod}. For instance, RSD~\cite{chen2021representation} addresses the challenge of feature scale changes during domain adaptation by aligning representation subspaces using a geometrical distance measure. Following a similar motivation, DARE-GRAM~\cite{Nejjar_2023_CVPR} aligns the angle and scale of the inverse Gram matrices for the source and target features. The authors address ill-conditioned inverse Gram matrices due to the ratio between batch size and the embedding dimension by proposing the use of the pseudo-inverse of the Gram matrix, achieved by selecting a specific subspace. Although traditional feature alignment methods can be applied in UDAR, their effectiveness is limited in this context. Regression models tend to learn focused representations, highly correlated with the target output, causing feature alignment to fail, as information about the domain shift in the input is not prominently present in the learned embeddings. This motivates the need to enrich the feature representation to guide feature alignment.

\subsection{Uncertainty-guided domain adaptation} Reducing prediction uncertainty on the target domain is a technique that has been explored for domain adaptation. The main intuition is that uncertainty for non-adapted target features is expected to be high, while it is expected to be low for source samples. In the setting of classification, entropy minimization and the related technique of self-training with pseudo-labels, effectively encourage cross-domain feature alignment~\cite{zou2018unsupervised,han2019unsupervised}. \cite{wen2019bayesian} proposed to minimize the uncertainty discrepancy between domains, where uncertainty is measured by the prediction entropy or variance using Monte-Carlo Dropout as an approximate Bayesian variational inference~\cite{gal2016dropout}. The approach proposed in~\cite{liang2019exploring} aligns the conditional distributions by minimizing the MMD between the mean embeddings of each class. The target distribution class mean embeddings are estimated by a sum of the elements weighted by their posterior probability. In the source-free DA setting,~\cite{roy2022uncertainty} proposed a method based on the information maximization loss but with uncertainty weighting of the samples, using a last-layer Laplace approximation to estimate the uncertainty of the pre-trained model.
While our work is closely related to~\cite{wen2019bayesian} which was applied to classification tasks, its extension to regression tasks has not been explored. Moreover, the existing studies focused on a narrow scenario, limited to Bayesian neural networks, particularly Monte Carlo Dropout, and adversarial alignment. This paper demonstrates that various uncertainty frameworks, including Gaussian Processes and evidential neural networks, can effectively guide alignment in regression tasks.

\section{Preliminaries}

\noindent \textbf{Deep Evidential Regression (DER).} To address the challenge of uncertainty quantification in UDAR, we adopt Deep Evidential Regression (DER), a method that simultaneously performs prediction and uncertainty estimation. Unlike more commonly applied uncertainty quantification regression methods for deep learning, DER  models both aleatoric (data-inherent) and epistemic (model-related) uncertainties.

From a maximum likelihood perspective, the likelihood of the parameters $\boldsymbol{\theta}$ given the $N$ observations $\mathbf{X}=\{\mathbf{x}_1, \dots, \mathbf{x}_{N} \}$ and regression targets $\mathbf{y}=\{y_1, \dots, y_{N} \}$ is defined as:
\begin{equation}
    p(\mathbf{y}|\mathbf{X},\boldsymbol{\theta}) = \prod\nolimits_{i=1}^{N} 
 p(y_i| \mathbf{x_i},\boldsymbol{\theta})
\end{equation}
In practice, it is commonly assumed that the target variable $y$ is drawn from a Gaussian distribution with mean $\mu = f(\mathbf{x})$ and variance $\sigma^2$. To effectively learn both aleatoric and epistemic uncertainties, the approach proposed by~\cite{amini2020deep} assumes that the mean and variance of the Gaussian distribution are themselves drawn from higher-order distributions: the mean from a Gaussian distribution and the variance from an Inverse-Gamma distribution. This hierarchical modeling choice allows for the incorporation of prior knowledge about the parameters. Specifically, the Normal Inverse-Gamma distribution with parameters ${(\gamma,\nu, \alpha, \beta)}$ is used as a conjugate prior:

\begin{equation}
y \sim \mathcal{N}(\mu, \sigma^2),\;
\mu \sim \mathcal{N}(\gamma, \sigma^2/\nu),\;
\sigma^2 \sim \Gamma^{-1}(\alpha, \beta),
\end{equation}

where $\gamma \in \mathbb{R}$ represents the predicted mean, $\nu > 0$ is the precision of the mean, which is inversely related to epistemic uncertainty, $\alpha > 1$ and $\beta > 0$ control the shape and scale of the inverse-gamma distribution, respectively, and are associated with aleatoric uncertainty. This formulation allows the model to quantify uncertainty in both its prediction ($\mu$) and the inherent noise in the data ($\sigma^2$).
The training process aims to minimize the negative log-likelihood (NLL) loss given by:

\begin{equation}
    \begin{aligned}
        \mathcal{L}_{\text{NLL}}= & \tfrac{1}{2} \log \left(\tfrac{\pi}{\nu}\right)-\alpha \log (2\beta(1+\nu)) +\log (\psi) 
        \\ 
        &+ \left(\alpha+\tfrac{1}{2}\right) \log \left((y-\gamma)^2 \nu+2\beta(1+\nu)\right)
    \end{aligned} 
\end{equation}

where $\psi = \Gamma(\alpha) /\Gamma(\alpha+1/2)$. The total evidential loss is defined as the sum of the negative log-likelihood loss and a regularization term: 
\vspace{-0.2cm}
\begin{equation}
    \mathcal{L_\text{EVI}}=\mathcal{L}_{\text{NLL}} + \lambda_{\text{EVI}}\mathcal{L}_R,
\end{equation}

where $\mathcal{L}_R = | y -\gamma| (2\nu+\alpha)$ penalizes evidence of prediction errors, and the hyperparameter $\lambda_{\text{EVI}}$  balances the trade-off between fitting the model to the data and increasing the uncertainty measures. 

In practice, this loss is minimized using standard gradient descent methods. This allows the neural network to learn not only point estimates for predictions but also full probability distributions,  capturing the uncertainty associated with each prediction.

\section{Methods}\label{sec11}

\subsection{Problem setting}

Let $\mathcal{X}$ and $\mathcal{Y}$ denote the input space and output spaces, respectively. We define the source domain  $S$ as consisting of $N_S$ labeled samples $\{(\mathbf{x}_i^S, y_i^S)\}_{i=1}^{N_S}$ drawn from the joint distribution $P_S$. Conversely, the target domain as $T$ comprises $N_T$ unlabeled samples $\{\mathbf{x}_i^T\}_{i=1}^{N_T}$ drawn from the from marginal distribution $P_T$. 
Our primary objective is to predict the continuous-valued outputs $y^T$ for unlabeled samples in the target domain by leveraging the labeled data from the source domain. To achieve this, we develop a model $f(\cdot)$ composed of two neural networks: a feature extractor $g(\cdot)$ that transforms the input data into meaningful feature representations, and a regression head $r(\cdot)$ that maps these extracted features to the continuous output space. Our objective is to train $f: \mathbf{x} \rightarrow y$, parameterized by $\mathbf{\theta}$, to minimize the expected error on the target data using the $\ell_2$ loss function. We assume that the labels in both domains cover the same range.

\subsection{Theoretical Motivation}

In Unsupervised Domain Adaptation for Regression (UDAR), a common approach is to align features between the source and target domains. However, perfect feature alignment can lead to suboptimal performance, as it may eliminate essential domain-specific information. To address this challenge, we present a theoretical framework that quantifies this limitation and motivates our uncertainty-guided approach. Central to this framework is the concept of the adaptation gap \cite{jiang2023understanding,dong2024simmmdg}, which measures  the difference in predictive power between the source and target domains:

\begin{definition}[Adaptation Gap]
Let $\mathbf{z}^S, y^S$ and  $\mathbf{z}^T, y^T$ denote the features extracted by the feature extractor $g(\cdot)$ and the labels for the source and the target domains, respectively. The adaptation gap $\Delta_p$ is defined as:

\begin{equation}
\Delta_p := I(\mathbf{z}^S; y^S) - I(\mathbf{z}^T; y^T)
\end{equation}

where $I(\mathbf{z}; y)$ represents the mutual information between $\mathbf{z}$ and $y$. We assume that $I(\mathbf{z}^S; y^S) \geq I(\mathbf{z}^T;y^T)$, ensuring $\Delta_p \geq 0$, given that the feature extractor has been trained on the source domain, involving a loss of information. In other words, the features extracted by $g(\cdot)$ provide less information about $y$ in the target domain than in the original source domain.
\end{definition}

\begin{theorem}
Let $g(\cdot)$ be a deterministic feature extractor, and let $\mathbf{z}^S = g(x^S)$, $\mathbf{z}^T = g(x^T)$ denote the extracted features for the source and target domains, respectively. Assuming that $y^S|\mathbf{z}^S$ and $y^T|\mathbf{z}^T$ follow  Gaussian distributions, if the features are perfectly aligned, i.e., $\mathbf{z}^S = \mathbf{z}^T = \mathbf{z}$, then:

\begin{align}
    \inf_f \mathbb{E}_{p_T}[\ell(f(\mathbf{z}^T), y^T)] 
    &- \inf_{f'} \mathbb{E}_{p_S}[\ell(f'(\mathbf{z}^S), y^S)] \nonumber \\
    &\geq \sigma_{y^S|x^S}^2 (e^{2\Delta_p} - 1)
\end{align}

where $\ell$ denotes  the mean squared error loss, $p_S$ and $p_T$ are the source and target domain distributions respectively, and $\sigma_{y^S|x^S}^2$ is the conditional variance of $y^S$ given $x^S$.
\end{theorem}

\vspace{6pt}

Our theorem highlights a fundamental limitation of perfect feature alignment in UDAR: as the adaptation gap $\Delta_p$ increases, the lower bound on expected loss grows exponentially. This result indicates a need for an approach that balances feature alignment with the preservation of domain-specific information. This finding motivates our proposed Uncertainty-Guided Alignment approach.

As shown in Figure~\ref{fig:density_sub}, models trained to predict uncertainty using evidential learning can distinguish between source and target domains in a single forward pass, while retaining domain-specific information. UGA leverages this capability to guide the alignment process by: (1) selectively aligning features based on uncertainty estimates, ensuring that domain-specific information is preserved where necessary; (2) incorporating uncertainty as a regularization term, encouraging the model to learn more robust and transferable features.  In addition, UGA is providing a mechanism to express confidence in the model's predictions.

\subsection{Overview of the proposed method}

\begin{figure*}[ht]
    \centering
    \includegraphics[width=0.9\textwidth]{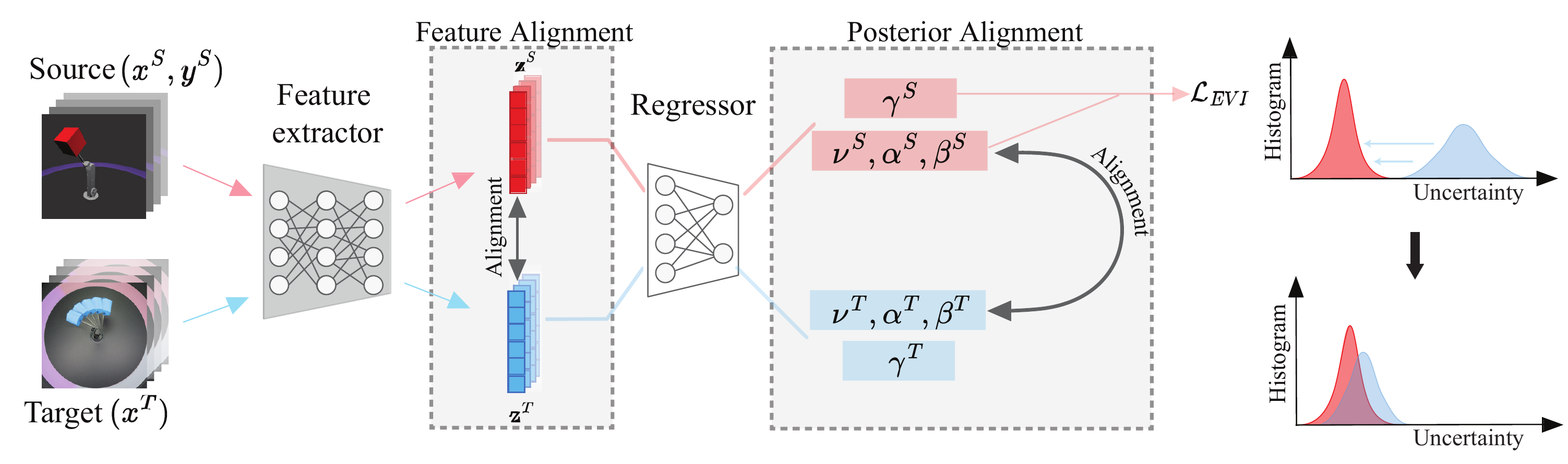}
    \caption{Overview of our proposed Uncertainty-Guided Alignment (UGA) framework for unsupervised domain adaptation in regression. UGA leverages the Deep Evidential Regression (DER) uncertainty framework. We propose aligning augmented feature representations with uncertainty to guide traditional feature alignment. We also introduce Posterior alignment, an approximation method aligning the evidential parameters.}
    \label{fig:main_figure}
\end{figure*}

Our proposed method, Uncertainty Guided Adaptation (UGA), addresses the challenge of unsupervised domain adaptation in regression by leveraging uncertainty to guide the alignment of source and target domain distributions. UGA introduces two core strategies: Uncertainty-Guided Feature Alignment and Uncertainty-Guided Posterior Alignment.

In the feature alignment approach, we align the distributions of feature embeddings between source and target domains, considering both the predicted values and their associated uncertainties. This ensures a comprehensive alignment that accounts for both regression outputs and the model’s confidence in those predictions.
Alternatively, the posterior alignment approach directly aligns the uncertainty distributions between domains at the posterior level. This serves as an approximation of feature alignment, focusing on matching uncertainty estimates across domains.

While our framework is flexible and can be combined with various feature alignment techniques and uncertainty quantification methods, we demonstrate it using a combination of Maximum Mean Discrepancy (MMD) and Deep Evidential Regression (DER) for simplicity and consistency. Figure~\ref{fig:main_figure} provides an overview of the proposed framework.

This choice is motivated by the fact that Bayesian neural networks (BNNs), traditionally a standard for uncertainty quantification treating model weights as distributions, are computationally expensive due to the requirement of multiple input samples to estimate output variance. To overcome this limitation, we adopt evidential deep learning, which frames learning as an evidence-acquisition process. This approach uses a standard neural network architecture,  significantly improving computational efficiency compared to  BNNs by eliminating the need for sampling. By leveraging evidential learning, we can estimate uncertainties in a single forward pass, enabling more efficient and scalable uncertainty-guided adaptation.

\subsubsection{Uncertainty-Guided Feature Alignment}

Uncertainty-guided Feature Alignment aligns the embeddings of source and target domains, following a similar approach to~\cite{long2015learning}, but with the addition of uncertainty guidance. Let the feature embeddings for the source and target domains be denoted as $\mathbf{z}_i^S=g(\mathbf{x}_i^S)$ and $\mathbf{z}_i^T=g(\mathbf{x}_i^T)$, respectively. To quantify  the discrepancy between these distributions, we compute the Maximum Mean Discrepancy (MMD) as:

\begin{equation}\label{feature}
\text{MMD}^2_{\text{feature}} = \bigl\rvert\bigl\rvert \frac{1}{N_S}\sum_{i=1}^{N_S}{\phi(\mathbf{z}_i^S)}-\frac{1}{N_T}\sum_{i=1}^{N_T} \phi(\mathbf{z}_i^T)~\bigl\lvert\bigl\lvert^2_{\mathcal{H}_k}.
\end{equation}

Here, the feature embeddings $z_i$ incorporate both the predicted target values and their associated uncertainties. By leveraging this uncertainty information, we can more effectively guide the minimization of feature discrepancies between the source and target domains. The final objective function for this alignment method is formulated as:

\begin{equation}
\mathcal{L_\text{feature}} = \mathcal{L_\text{EVI}} + \lambda \cdot \text{MMD}^2_{\text{feature}}
\label{eq:feature}
\end{equation}

where $\lambda$ is a hyperparameter that controls the relative weight of the feature alignment term, and $\mathcal{L_\text{EVI}}$ represents the supervised loss of the source domain. 

\subsubsection{Uncertainty-Guided Posterior Alignment}

The Uncertainty-Guided Posterior Alignment method takes a different approach by directly aligning the uncertainty distributions between the source and target domains. Unlike feature alignment, which focuses on aligning feature representations, this method focuses on aligning the uncertainty estimates.

To achieve this, we compute the MMD loss between the higher-order parameters of the uncertainty distributions produced by the DER model. The evidential network outputs for the source and target domains are denoted as $\gamma_i^S, \nu_i^S, \alpha_i^S, \beta_i^S = f(\textbf{x}_i^S)$ and $\gamma_i^T, \nu_i^T, \alpha_i^T, \beta_i^T = f(\textbf{x}_i^T)$, respectively. In this context, $\gamma$ represents the predicted mean, while $\nu$, $\alpha$, and $\beta$ parameterize the uncertainty distribution. Alignment is achieved by minimizing the squared MMD loss:

\begin{equation}
    \begin{aligned}
\label{posterior}
\text{MMD}^2_{\text{posterior}} = &\bigl\rvert\bigl\rvert \frac{1}{N_S}\sum_{i=1}^{N_S}{\phi([\nu_i^S, \alpha_i^S, \beta_i^S])}\\  
&-\frac{1}{N_T}\sum_{i=1}^{N_T} \phi([\nu_i^T, \alpha_i^T, \beta_i^T])~\bigl\lvert\bigl\lvert^2_{\mathcal{H}_k}.
 \end{aligned}
\end{equation}
   
This alignment ensures that the uncertainty estimates captured by the evidential network, represented by the parameters $\nu$, $\alpha$, and $\beta$, remain consistent and transferable across different domains. The final objective function to minimize is defined as follows:

\begin{equation}
\mathcal{L_\text{posterior}}= \mathcal{L_\text{EVI}} + \lambda \cdot \text{MMD}^2_{\text{posterior}}.
\label{eq:posterior}
\end{equation}

\section{Experiments}

We evaluate the performance of our proposed method using two computer vision benchmarks and a Prognostic and Health Management (PHM) task focused on battery State of Charge prediction. This section details our experimental setup, datasets, and evaluation metrics utilized.

\subsection{Datasets and Tasks}

\subsubsection{Computer Vision Benchmarks\\}

\begin{figure}[h!]
\centering
\begin{minipage}[b]{0.25\textwidth}
    \centering
    \includegraphics[width=\textwidth]{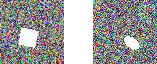}
    \caption*{(a) Noise}
    \label{fig:noise}
\end{minipage}
\hfill
\begin{minipage}[b]{0.25\textwidth}
    \centering
    \includegraphics[width=\textwidth]{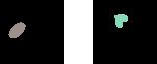}
    \caption*{(b) Color}
    \label{fig:Color}
\end{minipage}
\hfill
\begin{minipage}[b]{0.25\textwidth}
    \centering
    \includegraphics[width=\textwidth]{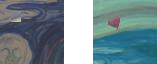}
    \caption*{(c) Scream}
    \label{fig:Scream}
\end{minipage}
\caption{Sample example of different domains in dSprites. (a) Noise domain. (b) Color domain. (c) Scream domain.}
\label{fig:Dsprites}
\end{figure}

\noindent\textbf{dSprites}~\cite{dsprites17} is a synthetic 2D image dataset with five independent latent factors. We use three variants as separate domains: Color (\textbf{C}), Scream (\textbf{S}), and Noise (\textbf{N}), as illustrated in Figure~\ref{fig:Dsprites}. Each domain consists of 737,280 images. Our tasks focus on regression for scale and position.

\begin{figure}[h!]
\centering
\begin{minipage}[b]{0.25\textwidth}
    \centering
    \includegraphics[width=\textwidth]{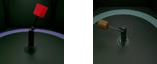}
    \caption*{(a) Real}
    \label{fig:real}
\end{minipage}
\hfill
\begin{minipage}[b]{0.25\textwidth}
    \centering
    \includegraphics[width=\textwidth]{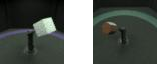}
    \caption*{(b) Realistic}
    \label{fig:RealistiC}
\end{minipage}
\hfill
\begin{minipage}[b]{0.25\textwidth}
    \centering
    \includegraphics[width=\textwidth]{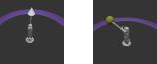}
    \caption*{(c) Toy}
    \label{fig:Toy}
\end{minipage}
\caption{Sample example of different domains in MPI3D. (a) Real domain. (b) Realistic domain. (c) Toy domain.}
\label{fig:MPI3D_datast}
\end{figure}

\noindent\textbf{MPI3D}~\cite{NEURIPS2019_d97d404b} contains 1,036,800 examples of 3D objects from three domains: Toy (\textbf{T}), RealistiC (\textbf{RC}), and ReaL (\textbf{RL}). This dataset enables the study of domain gaps between real and simulated data in a robotics context. We focus on regression tasks involving rotation around vertical and horizontal axes. Figure~\ref{fig:MPI3D_datast} showcases sample examples from each domain, highlighting the variations in appearance and characteristics.

\subsubsection{PHM task: State-of-Charge prediction}

\begin{figure}[h!]
    \centering
    \includegraphics[width=0.49\textwidth]{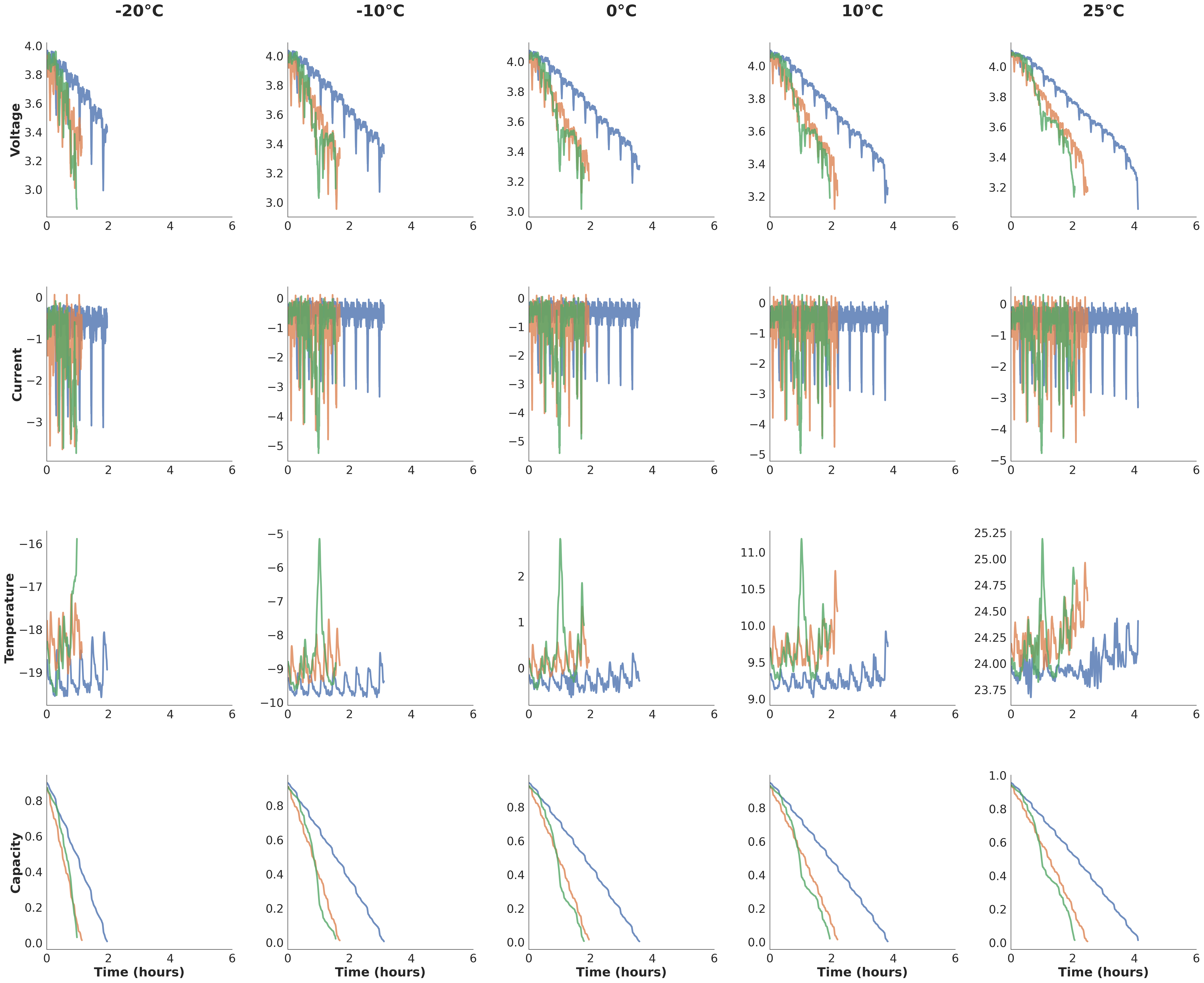}
    \caption{Example of State of Charge (SoC) curves and associated features (voltage, current, temperature, and time) for LG batteries that have been charged and discharged until reaching 0\% capacity. The plots illustrate the impact of temperature on battery capacity, discharge rate, and overall performance throughout the battery's lifecycle.}
    \label{fig:example_lg}
\end{figure}

\begin{figure}[h!]
    \centering
    \includegraphics[width=0.49\textwidth]{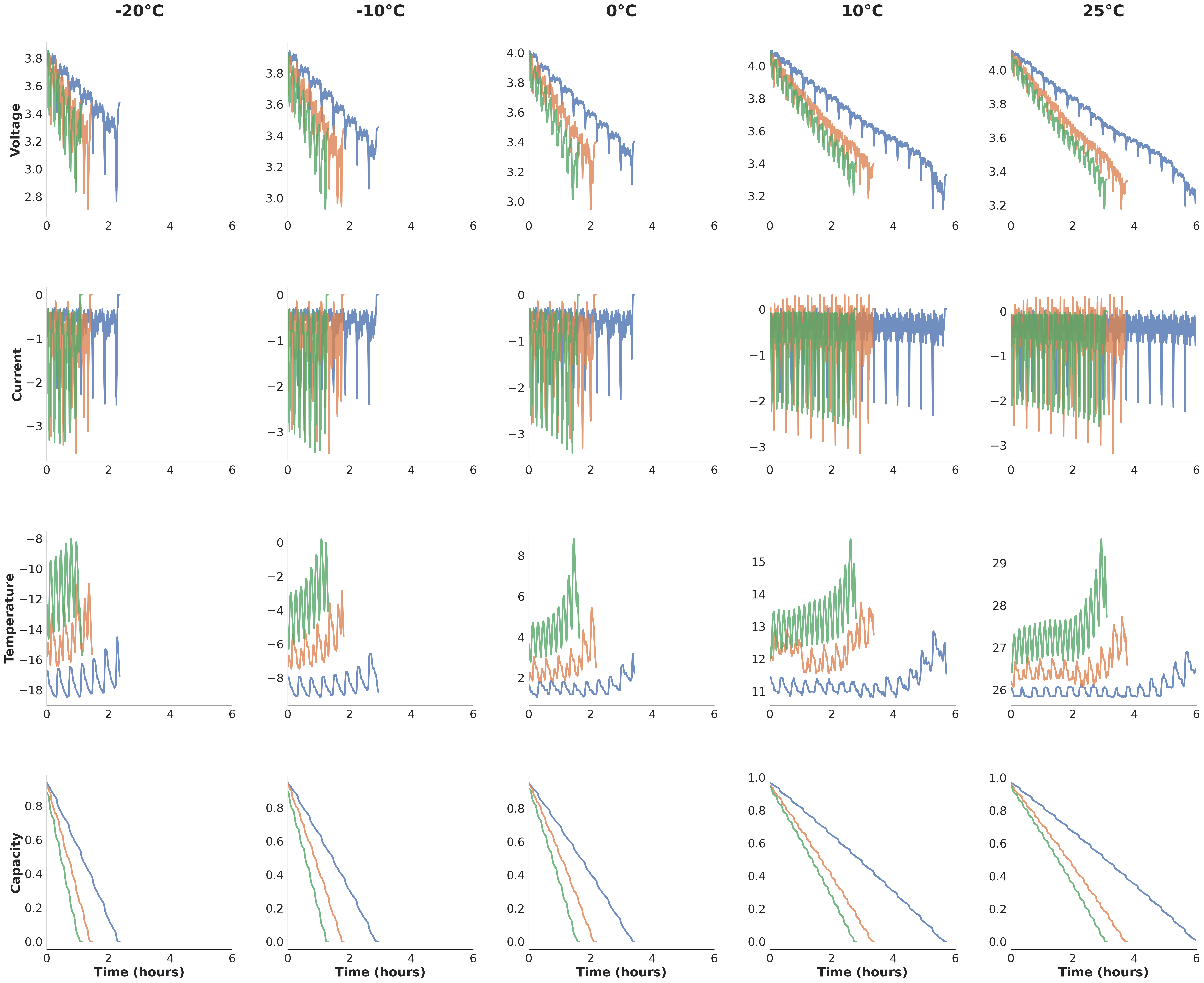}
    \caption{Comparison of SoC curves and associated features (voltage, current, temperature, and time) for Panasonic batteries that have been charged and discharged until reaching 0\% capacity. The plots highlight the differences in battery performance between the two manufacturers, including variations in capacity, discharge rate, and temperature sensitivity throughout the battery's lifecycle.}
    \label{fig:example_panasonic}
    \vspace{-0.5cm}
\end{figure}

For the state-of-charge (SOC) prediction task, we use two publicly available datasets:
\begin{itemize}
    \item LG 18650HG2 dataset~\cite{kollmeyer2020lg}, examples are seen in Figure~\ref{fig:example_lg}.
    \item Panasonic 18650PF dataset~\cite{kollmeyer2018panasonic},examples are seen in Figure~\ref{fig:example_panasonic}.
\end{itemize}

The LG dataset captures the discharge behavior of a 3Ah LG HG2 cell tested at temperatures ranging from -20°C to 40°C, while the Panasonic dataset uses a 2.9Ah 18650PF cell tested from -20°C to 25°C. Both datasets provide time-series measurements of voltage, current, battery case temperature, amp-hours, watt-hours, and power for each test under various standardized drive cycles. These drive cycles simulate different driving conditions and are summarized in Table \ref{tab:drive_cycles}. These datasets were collected specifically for State-of-Charge estimation in lithium-ion batteries. 

\begin{table}[h]
\centering
\fontsize{9pt}{10pt}\selectfont 
\setlength{\tabcolsep}{4pt} 
\setlength{\aboverulesep}{0pt}
\setlength{\belowrulesep}{0pt}
\setlength{\extrarowheight}{1pt} 
\caption{Description of Drive Cycles. NN: Neural Network cycle (Panasonic dataset only). Mixed: Random combinations of other cycles (LG dataset only).}
\begin{tabular}{ll}
\hline
Drive Cycle & Description \\
\hline
US06 & Aggressive, high speed and/or high acceleration \\
HWFET & Highway driving conditions under 60 mph \\
UDDS & City driving conditions with frequent stops \\
LA92 & More aggressive urban driving than UDDS \\
NN & Combination of US06 and LA92 \\
Mixed & Random combinations of above cycles \\
\hline
\end{tabular}
\label{tab:drive_cycles}
\end{table}

Our study focuses on adapting SOC prediction models across different lithium-ion battery types and operating conditions. We identify two domain gaps: 
\begin{enumerate}
    \item Manufacturer difference: LG and Panasonic batteries have unique chemical compositions and performance characteristics. 
    \item  Temperature variations: Battery behavior changes significantly based on temperature conditions.
\end{enumerate}

Our objective is to adapt a SOC prediction model trained on one battery type at a specific temperature to another battery type at a different temperature (eg. adapting a model trained on LG batteries at 25°C to Panasonic at 0°C). 
This task presents a challenging unsupervised domain adaptation problem as it involves simultaneously bridging both domain gaps (manufacturer differences and temperature variations).

During  data preprocessing, we downsample the dataset to 1Hz to reduce computational complexity while maintaining adequate temporal resolution. We select voltage, current, and temperature as input features, with SOC as the target variable. 

For our train-test split, we reserve specific drive cycles exclusively for testing and use the remaining cycles for training.  Specifically, for the Panasonic dataset, US06, LA92, and NN cycles are used for testing, while for the LG dataset, US06, LA92, and HWFET cycles serve as the test set. This partitioning ensures that our models are evaluated on a diverse range of driving conditions that were not seen during training.

\subsection{Experimental Setup}

\subsubsection{Computer Vision Experiments}

We use a pre-trained ResNet-18~\cite{https://doi.org/10.48550/arxiv.1512.03385} as the backbone, modified by disabling batch normalization and adding dropout (p=0.1) after each block. For DER, we use a single linear layer. We employ the Adam optimizer with a learning rate of $2e^{-5}$ and weight decay of $1e^{-3}$. The number of iterations is set to 20,000 for dSprites and 10,000 for MPI3D (2,000 for the S$\rightarrow$N task), following~\cite{chen2021representation} with a batch size of $b=36$. We normalize the source and target labels to a common range of $[0, 1]$. The results are reported using the Mean Absolute Error (MAE) as the primary evaluation metric, in line with previous studies~\cite{chen2021representation,Li_2021_CVPR}. 

\subsubsection{PHM Experiments}

For battery SOC prediction, we use a two-layer LSTM followed by a single linear layer. The model processes three inputs (Voltage, Current, and Temperature) provided as time series with a sequence length of 100. We set the hidden dimension to 64. Training is conducted using the SGD optimizer with a learning rate of $1e^{-1}$ for the final layers and $1e^{-2}$ for the LSTM, with weight decay of $1e^{-4}$ and a batch size of 128. All models are trained for 30 epochs. We report the Mean Square Error (MSE) as the primary metric and include coefficient of determination (R$^2$) for interpretability, given that the state of charge trajectory over time approximates a near-linear curve.

\subsection{Hyperparameters}

For both computer vision and PHM tasks, we set the hyperparameter $\lambda$ in Equations \ref{eq:feature} and \ref{eq:posterior} to increase from 0 to 1 during training, following $\lambda = \frac{2}{1+{\rm exp}(-10 \text{p})}-1$, where $\rm p$ represents the training progress. We set $\lambda_{\text{EVI}}$ to 1 for computer vision tasks and 0.1 for the PHM task.

\section{Results and Discussion}

\subsection{Computer Vision Results}

\begin{table}[h!]
\centering
\fontsize{7pt}{9pt}\selectfont 
\setlength{\tabcolsep}{2pt} 
\setlength{\aboverulesep}{0pt}
\setlength{\belowrulesep}{0pt}
\setlength{\extrarowheight}{1pt} 
\caption{Comparisons on dSprites regression tasks (sum of MAE across three targets). \xmark: no uncertainty (MSE loss), DER: Deep Evidential Regression. We compare JDOT, CORAL, and MMD with/without uncertainty-guided alignment. Bold results indicate best performance. }
\begin{tabular}{l|c|cccccc|c}
\toprule
Method & Unc. & C$\to$N & C$\to$S & N$\to$C & N$\to$S & S$\to$C & S$\to$N & Avg \\
\midrule
Source only & \xmark & 0.94 & 0.90 & 0.16 & 0.65 & 0.08 & 0.26 & 0.498 \\
DANN & \xmark & 0.47 & 0.46 & 0.16 & 0.65 & 0.05 & 0.10 & 0.315 \\
RSD & \xmark & 0.31 & 0.31 & 0.12 & 0.53 & 0.07 & 0.08 & 0.237 \\
DARE-GRAM & \xmark & 0.30 & 0.20 & 0.11 & 0.25 & 0.05 & 0.07 & 0.164 \\
\midrule
JDOT & \xmark & 0.86 & 0.79 & 0.19 & 0.64 & 0.10 & 0.23 & 0.468 \\
\textbf{UGA-Post. JDOT} & DER & 0.09 & 0.13 & 0.11 & 0.18 & 0.05 & 0.12 & 0.114 \\
\textbf{UGA-Feat. JDOT} & DER & 0.08 & 0.11 & 0.08 & 0.17 & 0.05 & 0.07 & 0.093 \\
\midrule
CORAL & \xmark & 0.95 & 0.87 & 0.13 & 0.57 & 0.05 & 0.12 & 0.448 \\
\textbf{UGA-Post. CORAL} & DER & 0.56 & 0.40 & 0.24 & 0.94 & 0.04 & 0.28 & 0.410 \\
\textbf{UGA-Feat. CORAL} & DER & 0.06 & 0.24 & 0.08 & 0.32 & \textbf{0.02} & 0.03 & 0.125 \\
\midrule
MMD & \xmark & 0.70 & 0.77 & 0.12 & 0.50 & 0.06 & 0.11 & 0.377 \\
\textbf{UGA-Post. MMD} & DER & \textbf{0.04} & \textbf{0.10} & 0.04 & \textbf{0.16} & 0.03 & 0.03 & 0.067 \\
\textbf{UGA-Feat. MMD} & DER & \textbf{0.04} & \textbf{0.10} & \textbf{0.02} & \textbf{0.16} & \textbf{0.02} & \textbf{0.02} & \textbf{0.060} \\
\bottomrule
\end{tabular}
\label{tab:results_dsprites}
\end{table}

\begin{table}[h!]
\centering
\fontsize{7pt}{9pt}\selectfont 
\setlength{\tabcolsep}{1pt} 
\setlength{\aboverulesep}{0pt}
\setlength{\belowrulesep}{0pt}
\setlength{\extrarowheight}{1pt} 
\caption{Comparisons on MPI3D regression tasks. Unc.: Uncertainty, \xmark: no uncertainty, DER: Deep Evidential Regression.}
\begin{tabular}{l|c|cccccc|c}
\toprule
Methods & Unc. & RL$\to$RC & RL$\to$T & RC$\to$RL & RC$\to$T & T$\to$RL & T$\to$RC & Avg \\
\midrule
Source only & \xmark & 0.17 & 0.44 & 0.19 & 0.45 & 0.51 & 0.50 & 0.38 \\

JDOT & \xmark  & 0.16 & 0.41 & 0.16 & 0.41 & 0.47 & 0.47 & 0.35 \\
DANN & \xmark & 0.09 & 0.24 & 0.11 & 0.41 & 0.48 & 0.37 & 0.28 \\
RSD & \xmark & 0.09 & 0.19 & 0.08 & 0.15 & 0.36 & 0.36 & 0.21 \\
DARE-GRAM & \xmark & 0.08 & 0.15 & 0.10 & 0.14 & \textbf{0.24} & \textbf{0.24} & \textbf{0.16} \\
\midrule
MMD & \xmark & 0.12 & 0.35 & 0.12 & 0.27 & 0.40 & 0.41 & 0.28 \\
\textbf{UGA-Post. MMD} & DER & 0.08 & 0.45 & 0.08 & 0.18 & 0.41 & 0.30 & 0.25 \\
\textbf{UGA-Feat. MMD} & DER & \textbf{0.07} & \textbf{0.13} & \textbf{0.07} & \textbf{0.11} & 0.34 & \textbf{0.24} & \textbf{0.16} \\
\bottomrule
\end{tabular}
\label{tab:results_MPI3D}
\end{table}

\noindent\textbf{dSprites:} Table~\ref{tab:results_dsprites} presents the results obtained on the dSprites dataset, comparing our proposed methods with existing domain adaptation approaches. Specifically, we consider DANN, RSD, and DARE-GRAM as state-of-the-art methods for UDAR. Additionally, we use three feature alignment methods JDOT, CORAL, and MMD. For each case, we report results obtained without uncertainty guidance and with the proposed frameworks: Feature alignment (UGA-Feature) and Posterior alignment (UGA-Posterior).

Our Uncertainty-Guided Alignment methods consistently outperform existing approaches across all transfer tasks. The UGA-Feature approach, in particular, shows substantial improvements when combined with various feature alignment methods. JDOT's average MAE improved from 0.468 to 0.093, CORAL's from 0.448 to 0.125, and MMD's from 0.377 to 0.060 when combined with UGA-Feature. These results significantly surpass the performance of state-of-the-art methods like RSD (0.237) and DARE-GRAM (0.164).
The UGA-Posterior approach also shows notable improvements, especially when combined with JDOT and MMD, achieving at least a 34\% relative improvement over RSD and 4\% over DARE-GRAM. Interestingly, while UGA-Posterior CORAL shows only a modest improvement from 0.448 to 0.410, UGA-Feature CORAL achieves a substantial gain to 0.125. This discrepancy might be attributed to CORAL's simplicity, which may not effectively align higher-order evidence in the posterior.

\noindent\textbf{MPI3D:} Table~\ref{tab:results_MPI3D} presents the results obtained on the MPI3D dataset.  Our proposed methods exhibit superior performance in five out of six tasks and achieve the best average performance. Across the six tasks, posterior alignment achieves an average MAE  of 0.250, while feature alignment achieves an average MAE of 0.160. However, in the T$\rightarrow$RL task, DARE-GRAM outperforms our proposed methods, with an MAE of 0.24, compared to  UGA-Feature's MAE of 0.34. In the appendix, we present the histogram of predicted values and observe that for the T$\rightarrow$RL task, the model struggles to predict values near the boundary (0 or 1).

\begin{table*}[h!]
\centering
\caption{Results for source LG and target Panasonic}
\fontsize{9pt}{11pt}\selectfont 
\setlength{\tabcolsep}{2pt} 
\setlength{\aboverulesep}{0pt}
\setlength{\belowrulesep}{0pt}
\setlength{\extrarowheight}{1pt} 
\begin{tabular}{l|*{6}{c@{\hspace{0.5em}}c|}c@{\hspace{0.5em}}c}
\hline
Transfer & \multicolumn{2}{c|}{Source only} & \multicolumn{2}{c|}{CORAL} & \multicolumn{2}{c|}{DAN} & \multicolumn{2}{c|}{JDOT} & \multicolumn{2}{c|}{DARE-GRAM} &  \multicolumn{2}{c|}{MMD} & \multicolumn{2}{c}{MMD+EVI} \\
Task & MSE & R² & MSE & R² & MSE & R² & MSE & R² & MSE & R² & MSE & R² & MSE & R² \\
\hline
-20 $\rightarrow$ -10 & 0.06 & 0.15 & 0.06 & 0.15 & 0.10 & -0.39 & 0.02 & 0.77 & 0.02 & 0.79 & 0.01 & 0.85 & 0.02 & 0.76 \\
-20 $\rightarrow$ 0 & 0.07 & 0.05 & 0.07 & 0.05 & 0.15 & -0.89 & 0.01 & 0.89 & 0.01 & 0.93 & 0.01 & 0.87 & 0.00 & 0.95 \\
-20 $\rightarrow$ 10 & 0.08 & -0.00 & 0.08 & -0.00 & 0.13 & -0.66 & 0.00 & 0.95 & 0.01 & 0.86 & 0.01 & 0.93 & 0.00 & 0.94 \\
-20 $\rightarrow$ 25 & 0.08 & -0.03 & 0.08 & -0.03 & 0.10 & -0.30 & 0.00 & 0.95 & 0.06 & 0.18 &  0.08 & -0.01 & 0.01 & 0.89 \\
\midrule
-10 $\rightarrow$ -20 & 0.10 & -0.33 & 0.10 & -0.32 & 0.06 & 0.18 & 0.04 & 0.50 & 0.04 & 0.41 &  0.03 & 0.59 & 0.03 & 0.55 \\
-10 $\rightarrow$ 0 & 0.05 & 0.31 & 0.05 & 0.31 & 0.07 & 0.06 & 0.01 & 0.88 & 0.01 & 0.91 &  0.01 & 0.87 & 0.01 & 0.94 \\
-10 $\rightarrow$ 10 & 0.08 & 0.00 & 0.08 & -0.00 & 0.11 & -0.39 & 0.01 & 0.93 & 0.00 & 0.94 & 0.01 & 0.90 & 0.00 & 0.95 \\
-10 $\rightarrow$ 25 & 0.09 & -0.11 & 0.09 & -0.16 & 0.08 & 0.01 & 0.01 & 0.86 & 0.01 & 0.92 & 0.08 & 0.03 & 0.00 & 0.98 \\
\midrule
0 $\rightarrow$ -20 & 0.06 & 0.16 & 0.06 & 0.16 & 0.08 & -0.13 & 0.04 & 0.51 & 0.05 & 0.34 & 0.04 & 0.41 & 0.05 & 0.39 \\
0 $\rightarrow$ -10 & 0.07 & 0.11 & 0.07 & 0.12 & 0.04 & 0.49 & 0.03 & 0.64 & 0.03 & 0.61  &  0.02 & 0.71 & 0.02 & 0.70 \\
0 $\rightarrow$ 10 & 0.04 & 0.46 & 0.04 & 0.48 & 0.08 & -0.08 & 0.00 & 0.95 & 0.01 & 0.92  & 0.00 & 0.94 & 0.00 & 0.96 \\
0 $\rightarrow$ 25 & 0.07 & 0.07 & 0.06 & 0.27 & 0.08 & -0.05 & 0.00 & 0.96 & 0.02 & 0.79& 0.00 & 0.94 & 0.00 & 0.97 \\
\midrule
10 $\rightarrow$ -20 & 0.16 & -1.13 & 0.12 & -0.66 & 0.13 & -0.79 & 0.04 & 0.47 & 0.07 & 0.10 & 0.10 & -0.32 & 0.04 & 0.42 \\
10 $\rightarrow$ -10 & 0.05 & 0.30 & 0.05 & 0.36 & 0.10 & -0.29 & 0.02 & 0.73 & 0.03 & 0.57 & 0.03 & 0.60 & 0.02 & 0.72 \\
10 $\rightarrow$ 0 & 0.02 & 0.72 & 0.02 & 0.73 & 0.04 & 0.43 & 0.01 & 0.82 & 0.02 & 0.78 & 0.01 & 0.83 & 0.01 & 0.87 \\
10 $\rightarrow$ 25 & 0.01 & 0.81 & 0.01 & 0.82 & 0.08 & -0.10 & 0.00 & 0.96 & 0.01 & 0.90 & 0.00 & 0.94 & 0.00 & 0.97 \\
\midrule
25 $\rightarrow$ -20 & 0.31 & -3.15 & 0.31 & -3.15 & 0.12 & -0.67 & 0.10 & -0.27 & 0.11 & -0.42 & 0.09 & -0.14 & 0.07 & 0.10 \\
25 $\rightarrow$ -10 & 0.29 & -2.83 & 0.29 & -2.83 & 0.12 & -0.65 & 0.04 & 0.40 & 0.07 & 0.03 & 0.08 & -0.06 & 0.03 & 0.56 \\
25 $\rightarrow$ 0 & 0.27 & -2.49 & 0.27 & -2.48 & 0.08 & -0.00 & 0.01 & 0.87 & 0.08 & -0.07 & 0.08 & -0.08 & 0.01 & 0.87 \\
25 $\rightarrow$ 10 & 0.21 & -1.72 & 0.20 & -1.63 & 0.10 & -0.27 & 0.01 & 0.86 & 0.01 & 0.88 & 0.01 & 0.88 & 0.01 & 0.92 \\
\hline
Average & 0.11 & -0.43 & 0.11 & -0.39 & 0.09 & -0.23 & \textbf{0.02} & \underline{0.73} & 0.03 & 0.57 & 0.04 & 0.53 & \textbf{0.02} & \textbf{0.77} \\
\hline
\end{tabular}
\label{tab:lg_to_panasonic}
\end{table*}

\begin{table*}[h!]
\centering

\caption{Results for source Panasonic and target LG}
\fontsize{9pt}{11pt}\selectfont 
\setlength{\tabcolsep}{2pt} 
\setlength{\aboverulesep}{0pt}
\setlength{\belowrulesep}{0pt}
\setlength{\extrarowheight}{1pt} 
\begin{tabular}{l|*{6}{c@{\hspace{0.5em}}c|}c@{\hspace{0.5em}}c}
\hline
Transfer & \multicolumn{2}{c|}{Source only} & \multicolumn{2}{c|}{CORAL} & \multicolumn{2}{c|}{DAN} & \multicolumn{2}{c|}{JDOT} & \multicolumn{2}{c|}{DARE-GRAM}  & \multicolumn{2}{c|}{MMD} & \multicolumn{2}{c}{MMD+EVI} \\
Task & MSE & R² & MSE & R² & MSE & R² & MSE & R² & MSE & R² & MSE & R² & MSE & R² \\
\hline
-20 $\rightarrow$ -10 & 0.03 & 0.47 & 0.03 & 0.47 & 0.02 & 0.69 & 0.03 & 0.54 & 0.03 & 0.53 & 0.04 & 0.43 & 0.02 & 0.62 \\
-20 $\rightarrow$ 0 & 0.02 & 0.72 & 0.02 & 0.72 & 0.01 & 0.87 & 0.02 & 0.72 & 0.02 & 0.77 & 0.02 & 0.74 & 0.01 & 0.82 \\
-20 $\rightarrow$ 10 & 0.01 & 0.79 & 0.01 & 0.79 & 0.01 & 0.79 & 0.02 & 0.71 & 0.02 & 0.75 & 0.02 & 0.77 & 0.01 & 0.83 \\
-20 $\rightarrow$ 25 & 0.05 & 0.33 & 0.05 & 0.33 & 0.06 & 0.08 & 0.02 & 0.71 & 0.02 & 0.75 & 0.01 & 0.85 & 0.02 & 0.77 \\
\midrule
-10 $\rightarrow$ -20 & 0.01 & 0.90 & 0.01 & 0.90 & 0.01 & 0.83 & 0.01 & 0.89 & 0.01 & 0.88 & 0.01 & 0.90 & 0.01 & 0.86 \\
-10 $\rightarrow$ 0 & 0.03 & 0.59 & 0.03 & 0.59 & 0.01 & 0.90 & 0.02 & 0.74 & 0.02 & 0.66 & 0.02 & 0.67 & 0.02 & 0.62 \\
-10 $\rightarrow$ 10 & 0.01 & 0.78 & 0.01 & 0.78 & 0.01 & 0.83 & 0.02 & 0.76 & 0.01 & 0.84 & 0.01 & 0.82 & 0.01 & 0.78 \\
-10 $\rightarrow$ 25 & 0.04 & 0.43 & 0.04 & 0.44 & 0.05 & 0.32 & 0.02 & 0.76 & 0.01 & 0.82 &  0.01 & 0.86 & 0.01 & 0.87 \\
\midrule
0 $\rightarrow$ -20 & 0.02 & 0.66 & 0.02 & 0.66 & 0.12 & -1.11 & 0.02 & 0.74 & 0.01 & 0.78 & 0.01 & 0.78 & 0.01 & 0.76 \\
0 $\rightarrow$ -10 & 0.01 & 0.83 & 0.01 & 0.83 & 0.02 & 0.76 & 0.01 & 0.91 & 0.00 & 0.92 & 0.01 & 0.92 & 0.01 & 0.88 \\
0 $\rightarrow$ 10 & 0.02 & 0.68 & 0.02 & 0.68 & 0.00 & 0.93 & 0.01 & 0.82 & 0.01 & 0.80 & 0.01 & 0.81 & 0.01 & 0.78 \\
0 $\rightarrow$ 25 & 0.02 & 0.69 & 0.02 & 0.69 & 0.16 & -1.35 & 0.01 & 0.82 & 0.01 & 0.86 &  0.01 & 0.92 & 0.01 & 0.91 \\
\midrule
10 $\rightarrow$ -20 & 0.20 & -2.44 & 0.20 & -2.43 & 0.09 & -0.62 & 0.02 & 0.62 & 0.02 & 0.63 &  0.04 & 0.28 & 0.03 & 0.50 \\
10 $\rightarrow$ -10 & 0.12 & -0.85 & 0.12 & -0.83 & 0.09 & -0.33 & 0.01 & 0.85  & 0.01 & 0.88 & 0.01 & 0.87 & 0.01 & 0.85 \\
10 $\rightarrow$ 0 & 0.05 & 0.29 & 0.05 & 0.30 & 0.05 & 0.29 & 0.00 & 0.94 
 & 0.00  & 0.96 & 0.00 & 0.94 & 0.00 & 0.94 \\
10 $\rightarrow$ 25 & 0.01 & 0.89 & 0.01 & 0.89 & 0.00 & 0.94 & 0.00 & 0.95 & 0.00 & 0.94 &  0.00 & 0.95 & 0.00 & 0.94 \\
\midrule
25 $\rightarrow$ -20 & 0.25 & -3.26 & 0.25 & -3.25 & 0.16 & -1.73 & 0.05 & 0.21 & 0.02 & 0.58 &  0.07 & -0.13 & 0.03 & 0.49 \\
25 $\rightarrow$ -10 & 0.22 & -2.32 & 0.22 & -2.32 & 0.16 & -1.50 & 0.02 & 0.76 & 0.03 & 0.52 & 0.02 & 0.65 & 0.01 & 0.82 \\
25 $\rightarrow$ 0 & 0.19 & -1.83 & 0.19 & -1.82 & 0.13 & -0.97 & 0.01 & 0.88 & 0.02 & 0.75 & 0.01 & 0.88 & 0.01 & 0.90 \\
25 $\rightarrow$ 10 & 0.12 & -0.84 & 0.12 & -0.79 & 0.08 & -0.20 & 0.01 & 0.92 & 0.00 & 0.97 &  0.00 & 0.93 & 0.00 & 0.93 \\
\hline
Average & 0.07 & -0.13 & 0.07 & -0.12 & 0.06 & 0.02 & 0.02 & 0.76 & \underline{0.01} & \underline{0.78}& 0.02 & 0.74 & \textbf{0.01} & \textbf{0.79} \\
\hline
\end{tabular}
\label{tab:panasonic_to_lg}
\end{table*}

\subsection{PHM Results}

Tables \ref{tab:lg_to_panasonic} and \ref{tab:panasonic_to_lg} present the results of the battery State of Charge prediction task, which involves adapting models between different battery manufacturers (LG to Panasonic and vice versa) and a range of operating temperatures (-20°C to 25°C). 

\noindent\textbf{LG to Panasonic Transfer:} In Table \ref{tab:lg_to_panasonic}, the source-only baseline model yields an average MSE of 0.11 and a negative R² of -0.43, highlighting the difficulty of adapting across different battery manufacturers and operating temperatures. Our proposed MMD+EVI approach significantly improves the SOC prediction performance, achieving an average MSE of 0.02 and an R² of 0.77. In contrast, the standard MMD method, which does not leverage uncertainty guidance, results in a higher MSE of 0.04 and a lower R² of 0.53. The incorporation of uncertainty guidance in MMD+EVI results in a 45\% relative improvement in R² compared to the standard MMD method. Other methods such as CORAL and DAN exhibit limited improvements over the baseline, with CORAL performing similarly to source-only (R² of -0.39) and DAN offering slight gains (R² of -0.23). JDOT shows strong alignment capabilities, closely following MMD+EVI with an R² of 0.73. DARE-GRAM also performs well, achieving an R² of 0.57, although it does not achieve the level of performance achieved by MMD+EVI.

\noindent\textbf{Panasonic to LG Transfer:} Table \ref{tab:panasonic_to_lg} presents the reverse adaptation scenario, where the baseline achieves a moderate performance with an MSE of 0.07 and an R² of -0.13, indicating a less severe domain shift compared to the LG to Panasonic direction. Our proposed MMD+EVI method excels in this setting, attaining the lowest MSE (0.01) and the highest R² of 0.79, which constitutes a 6.8\% relative improvement over the standard MMD method that achieves an R² of 0.74. JDOT remains competitive with an R² of 0.76, while DARE-GRAM closely follows with an R² of 0.78. Additionally, CORAL and DAN slightly outperform the baseline, achieving  R² values of -0.12 and 0.02, respectively.

\noindent\textbf{Highlights:} Our proposed MMD+EVI method excels in scenarios involving extreme temperature alternative  (e.g., shifting from 25°C to -20°C), as it handles these larger domain shifts more effectively than other approaches. For example, in the LG to Panasonic transfer at 25°C $\rightarrow$ -20°C, MMD+EVI achieves an R² of 0.10, outperforming standard MMD (-0.14), CORAL (-3.15), and DAN (-0.67). A similar trend is observed in the Panasonic to LG 25°C $\rightarrow$ -20°C transfer, where MMD+EVI achieves an R² of 0.49, with the standard MMD, CORAL, and DAN achieving -0.13, -3.25, and -1.73, respectively. These results underline the importance of handling extreme domain shifts and the effectiveness of uncertainty-guided adaptation in mitigating these challenges.

\begin{figure*}[h]
    \centering
    \begin{minipage}[b]{0.2\textwidth}
        \centering
        \includegraphics[width=\textwidth]{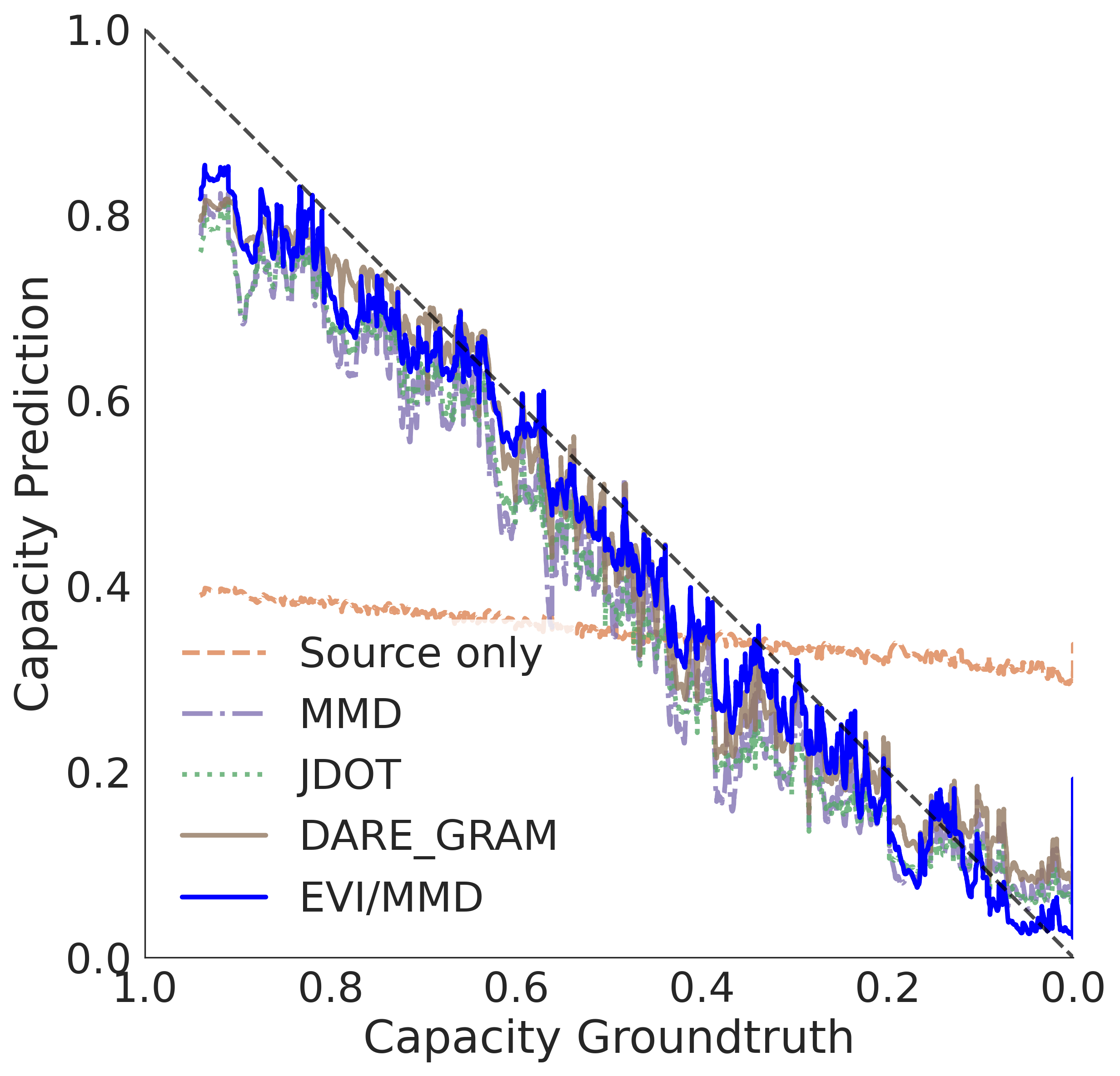}
        \caption*{(a) LG -20°C → Panasonic 0°C}
    \end{minipage}
    \hfill
    \begin{minipage}[b]{0.2\textwidth}
        \centering
        \includegraphics[width=\textwidth]{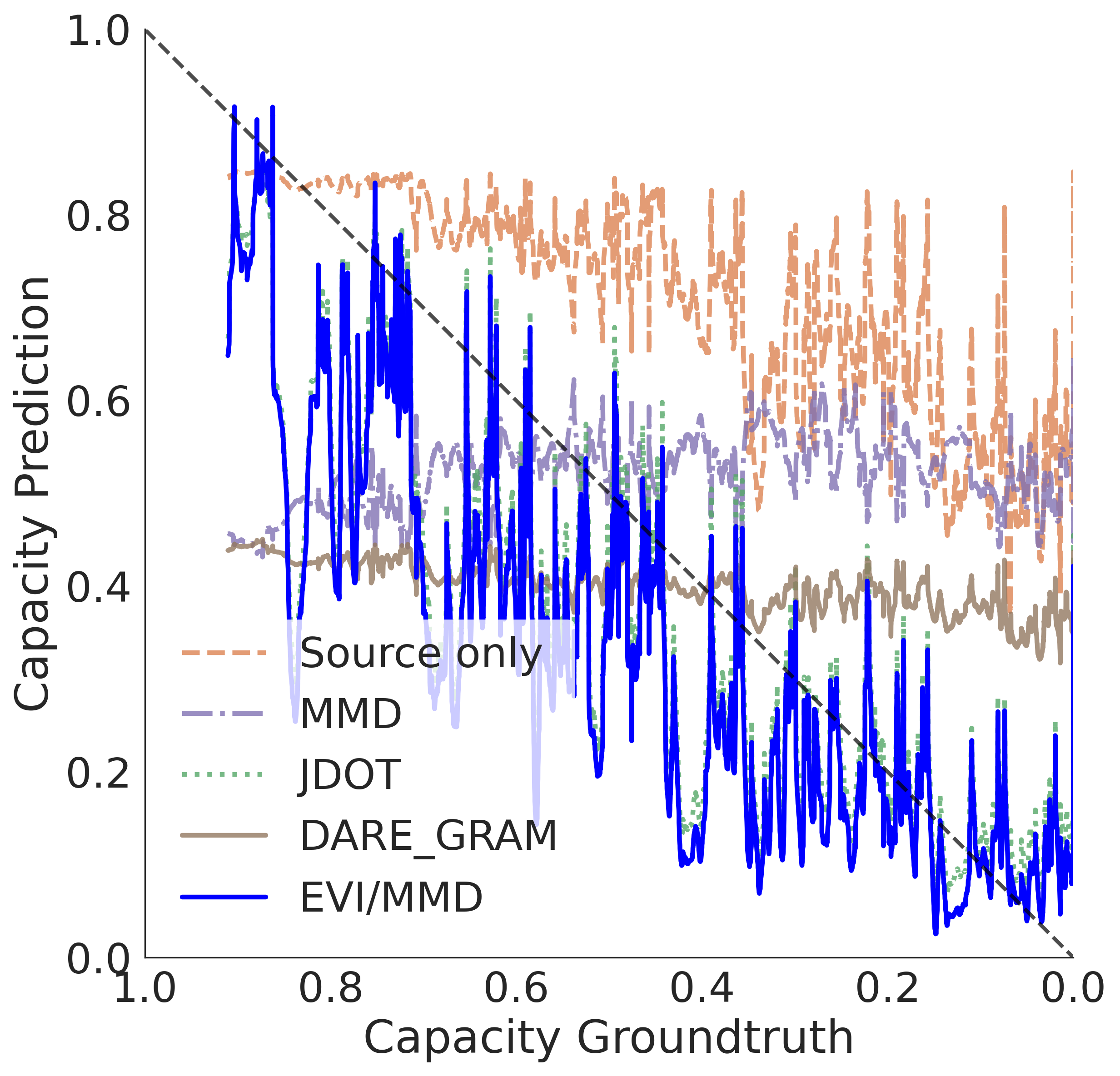}
        \caption*{(b) LG 10°C → Panasonic -20°C}
    \end{minipage}
    \hfill
    \begin{minipage}[b]{0.2\textwidth}
        \centering
        \includegraphics[width=\textwidth]{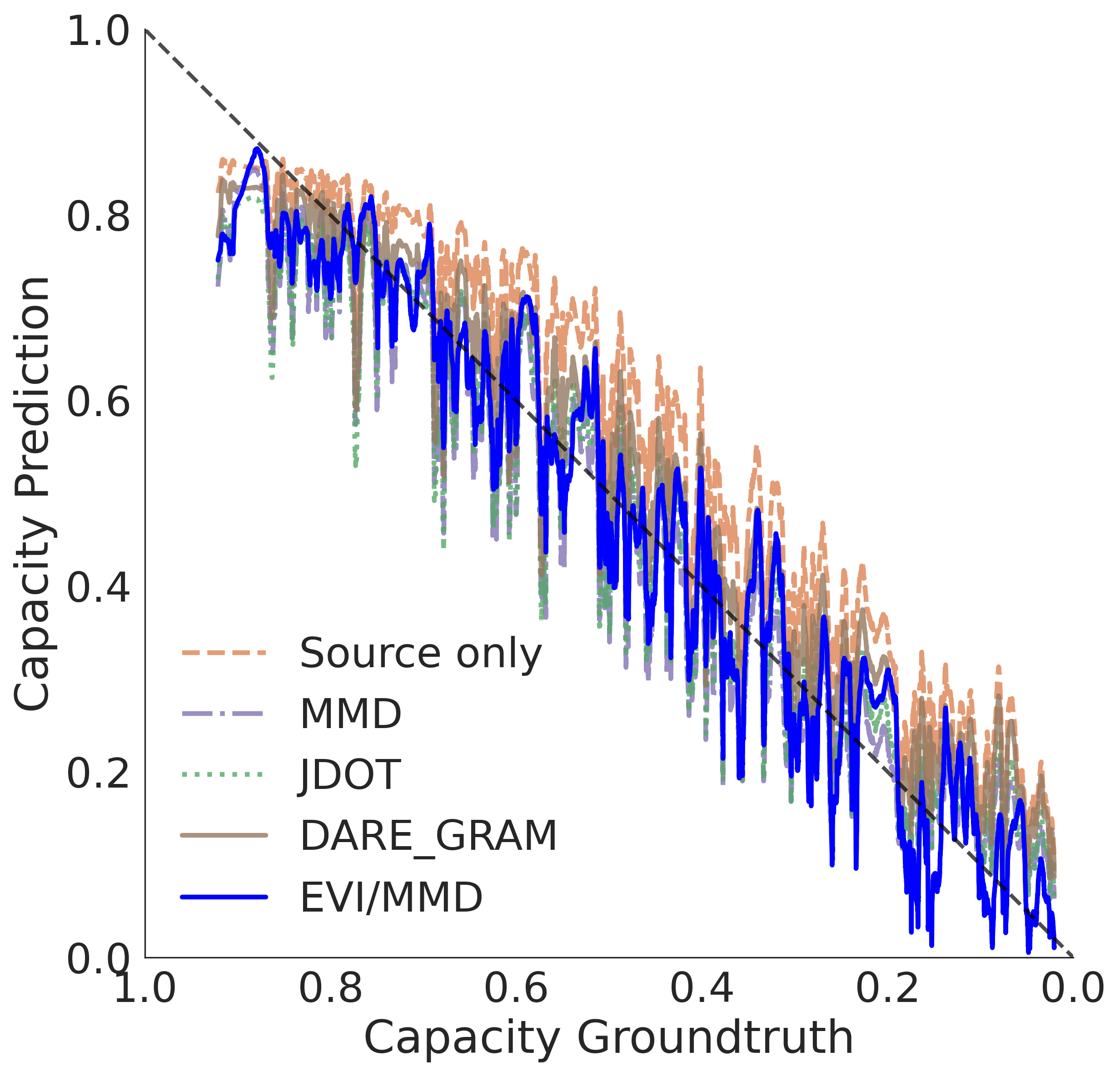}
        \caption*{(c) Panasonic -20°C → LG 0°C}
    \end{minipage}
    \hfill
    \begin{minipage}[b]{0.2\textwidth}
        \centering
        \includegraphics[width=\textwidth]{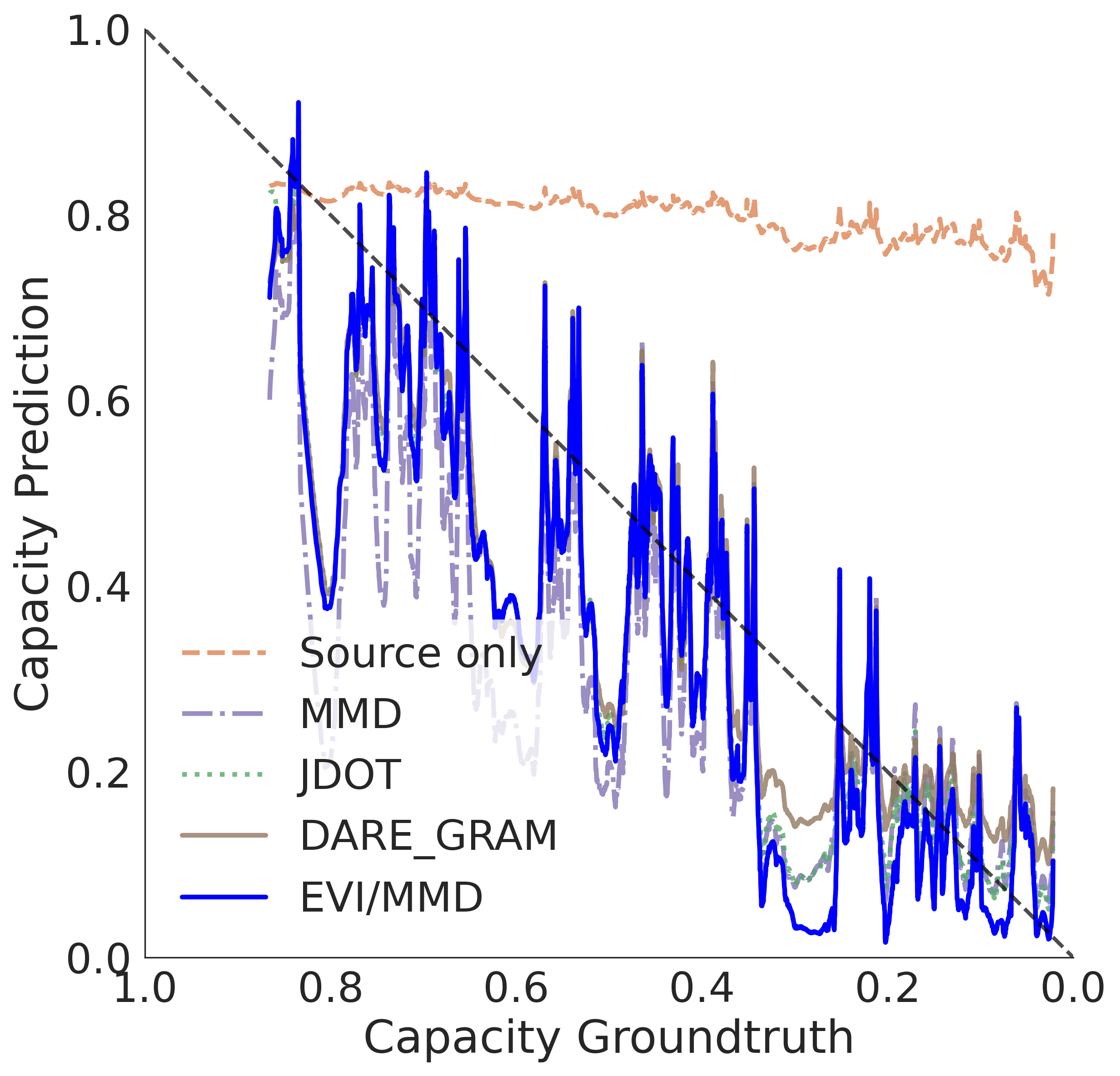}
        \caption*{(d) Panasonic 10°C → LG -20°C}
    \end{minipage}
    
    \vspace{1em}
    
    \begin{minipage}[t]{0.2\textwidth}
        \centering
        \includegraphics[width=\textwidth]{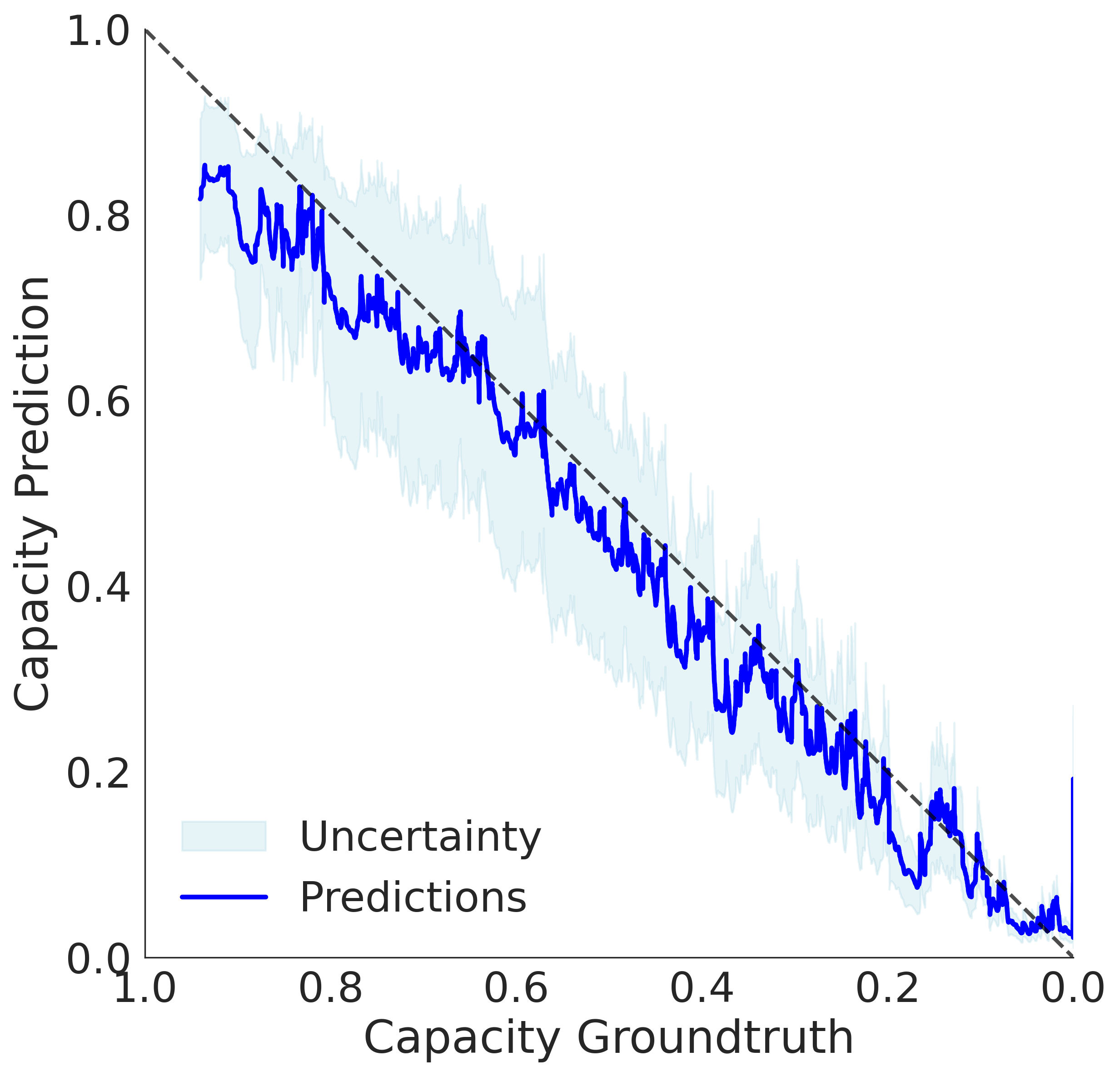}
        \caption*{(e) LG -20°C → Panasonic 0°C (MMD+EVI)}
    \end{minipage}
    \hfill
    \begin{minipage}[t]{0.2\textwidth}
        \centering
        \includegraphics[width=\textwidth]{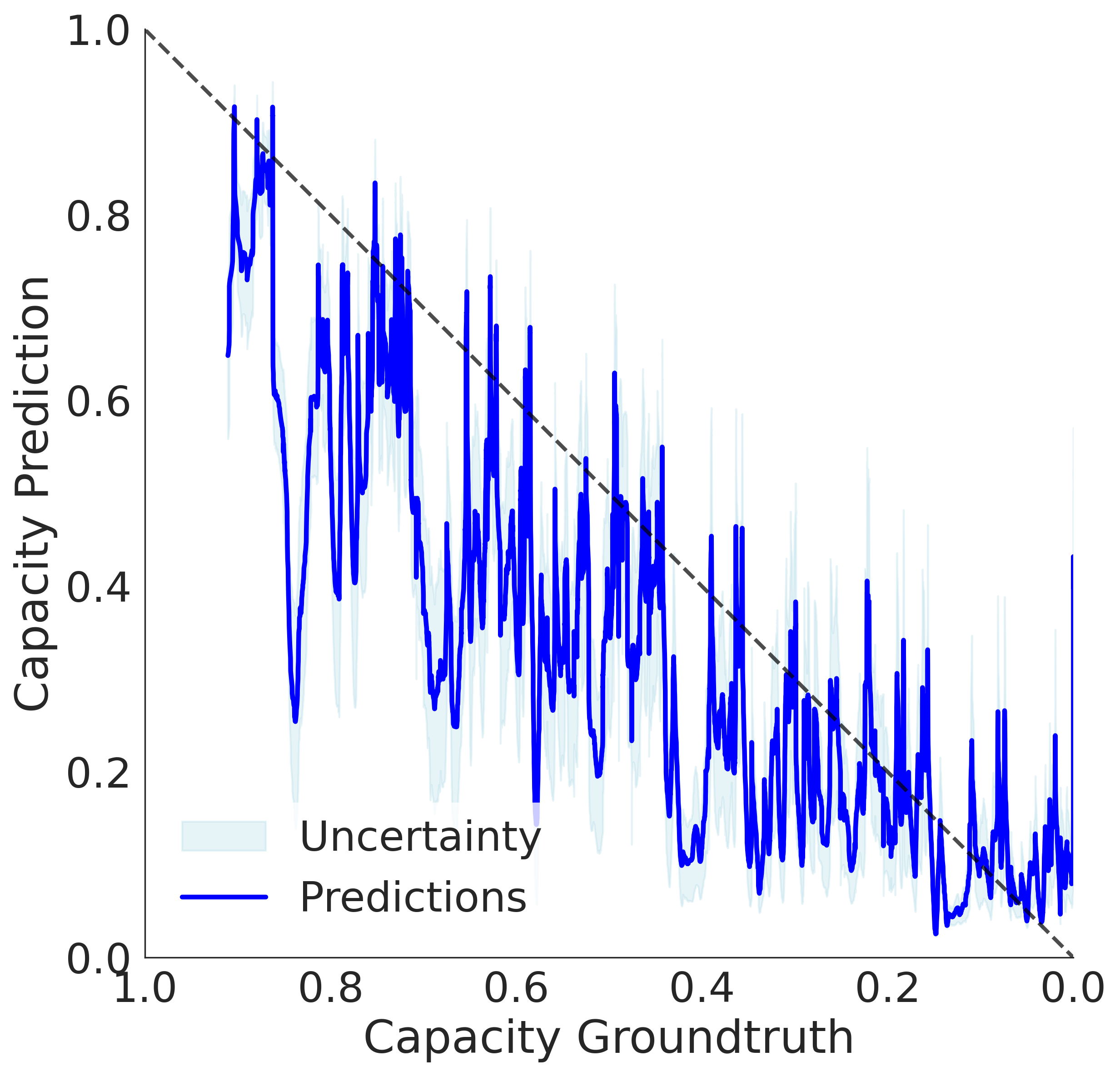}
        \caption*{(f) LG 10°C → Panasonic -20°C (MMD+EVI)}
    \end{minipage}
    \hfill
    \begin{minipage}[t]{0.2\textwidth}
        \centering
        \includegraphics[width=\textwidth]{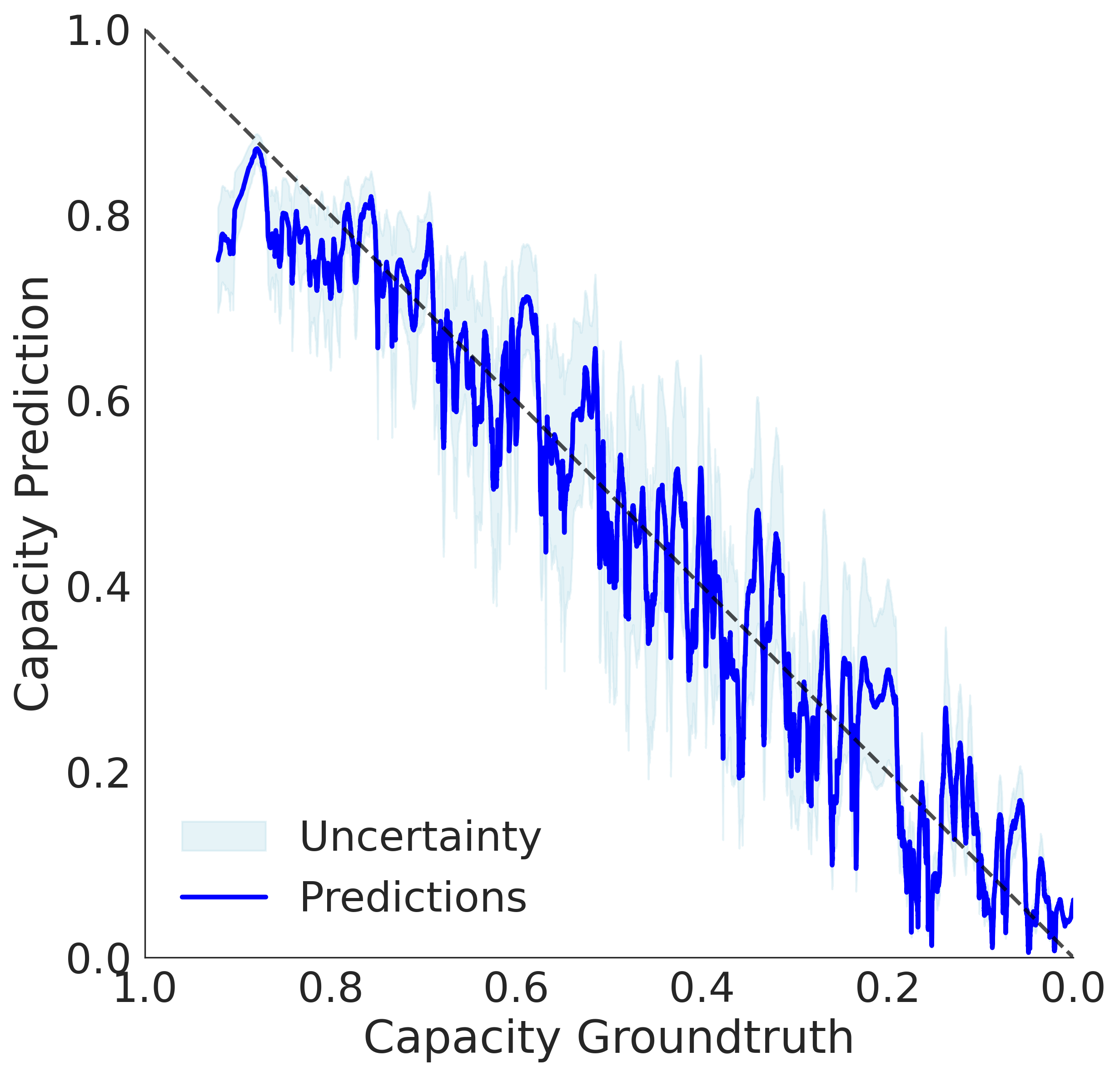}
        \caption*{(g) Panasonic -20°C → LG 0°C (MMD+EVI)}
    \end{minipage}
    \hfill
    \begin{minipage}[t]{0.2\textwidth}
        \centering
        \includegraphics[width=\textwidth]{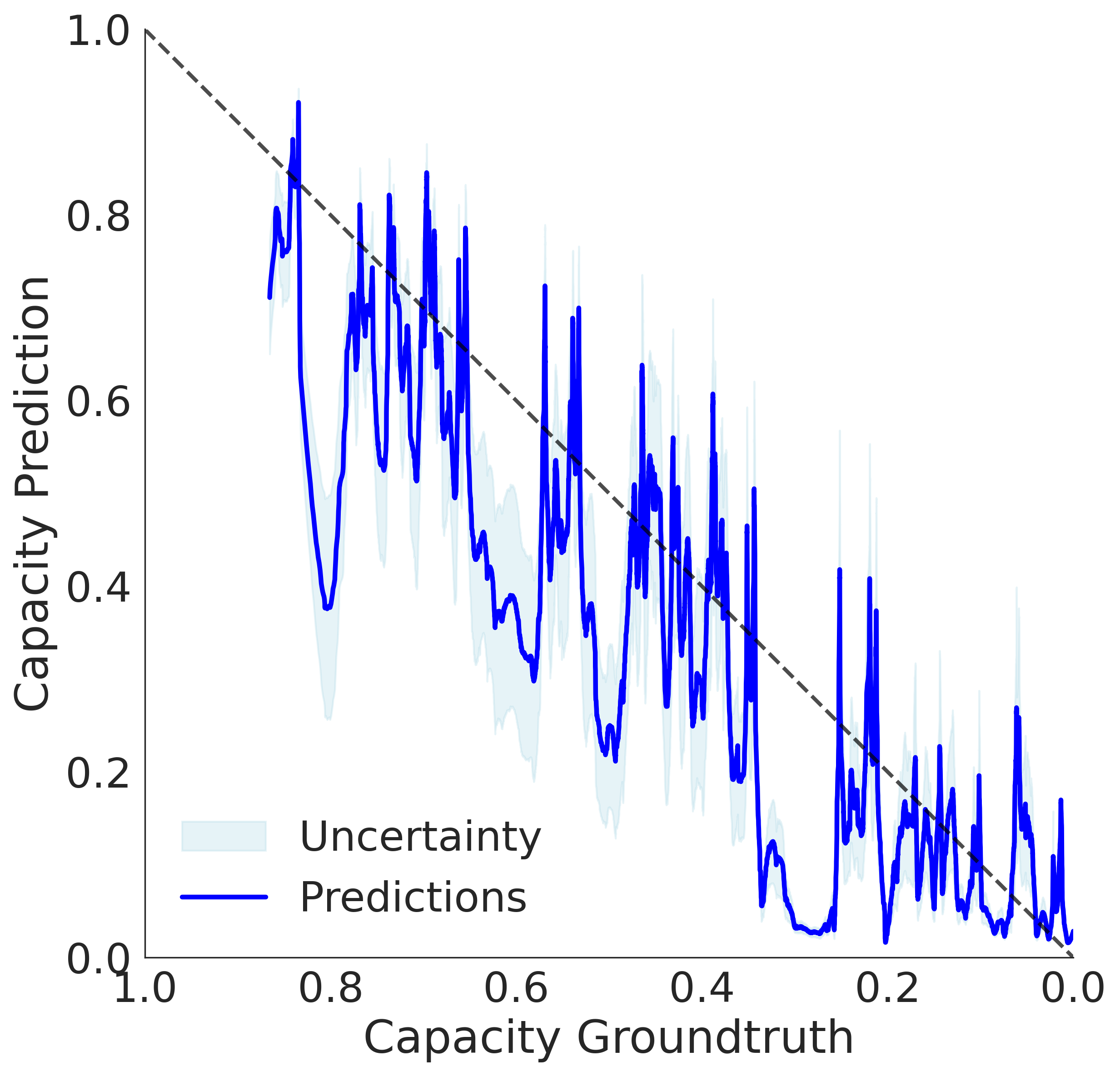}
        \caption*{(h) Panasonic 10°C → LG -20°C (MMD+EVI)}
    \end{minipage}
    
    \caption{Prediction trajectories for battery SoC across different transfer tasks. (a-d) Comparison of various methods. (e-h) MMD+EVI predictions with uncertainty estimates (shaded area represents $2\sigma$).}
    \label{fig:predictions_and_uncertainty}
\end{figure*}

\noindent\textbf{Qualitative results:} Figure \ref{fig:predictions_and_uncertainty} illustrates the qualitative performance of our proposed method and baselines on various transfer tasks in battery SoC prediction.

Figures \ref{fig:predictions_and_uncertainty}(a-d) illustrate  the prediction trajectories of different methods across various transfer tasks. Our proposed MMD+EVI approach  consistently  achieves closer alignment with the ground truth compared to other approaches, particularly in challenging scenarios involving large temperature differences (b and d).
Additionally, Figures \ref{fig:predictions_and_uncertainty}(e-h) showcase the MMD+EVI predictions with their associated uncertainty estimates. A clear trend is observed in the uncertainty behavior: it is initially  high at the start of the battery's lifecycle and progressively decreases as the battery ages. This pattern aligns with the expectation  that prediction confidence increases with the accumulation of more data  over the battery's lifespan, enhancing the model's reliability over time.

\begin{figure*}[h!]
\centering
\begin{minipage}[t]{0.4\textwidth}
    \centering
    \includegraphics[width=\textwidth]{new_figures_battery/lg_-20_uncertainty_comparison_None.png}
    \caption*{(a) DER without alignment}
    \label{fig:no_box}
\end{minipage}
\hspace{20pt}
\begin{minipage}[t]{0.4\textwidth}
    \centering
    \includegraphics[width=\textwidth]{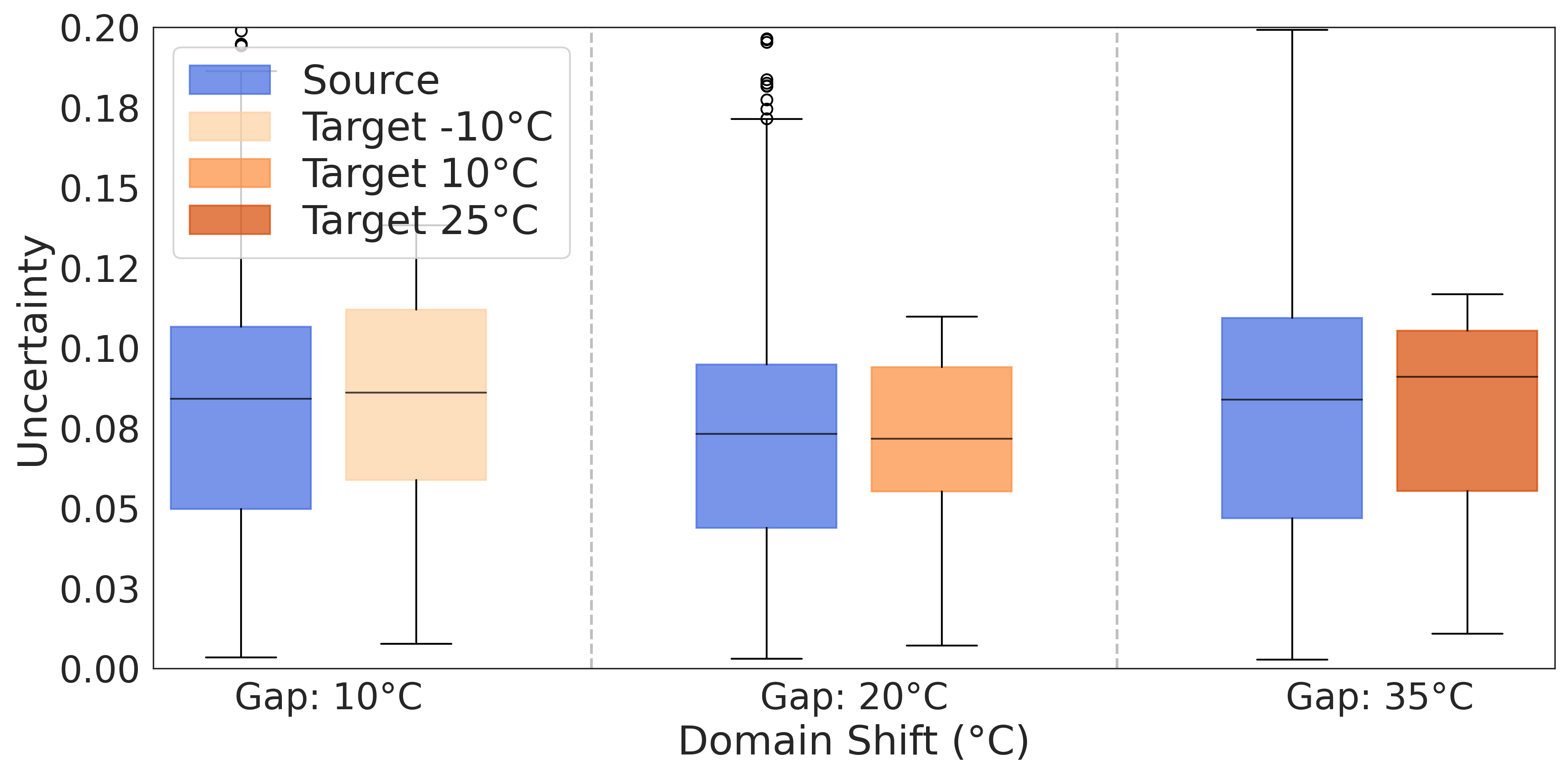}
    \caption*{(b) UGA-Feature alignment}
    \label{fig:feature_box}
\end{minipage}
\caption{Uncertainty distribution comparison between source (-20°C) and target domains of increasing temperature for LG batteries, (a) without alignment and (b) with UGA-Feature alignment.}
\label{fig:uncertainty_box}
\end{figure*}

The effectiveness of our uncertainty-guided approach is further illustrated in Figure \ref{fig:uncertainty_box}, which compares uncertainty distributions between source and target domains.

Figure \ref{fig:uncertainty_box}(a) shows a significant discrepancy in uncertainty distributions between source and target domains when no alignment is applied. In contrast, Figure \ref{fig:uncertainty_box}(b) demonstrates that our UGA-Feature alignment method effectively minimizes this discrepancy, resulting in highly comparable uncertainty distributions across both domains. This alignment indicates that our proposed approach successfully calibrates the model's confidence under varying operating conditions, thereby enhancing its cross-domain performance.

\section{Ablation study} 

To further investigate the impact of different uncertainty quantification methods within our approach, we conducted an ablation study on the dSprites dataset, as shown in Table \ref{tab:gp_vs_der}.

\begin{table}[h!]
\centering
\fontsize{7pt}{9pt}\selectfont 
\setlength{\tabcolsep}{1pt} 
\setlength{\aboverulesep}{0pt}
\setlength{\belowrulesep}{0pt}
\setlength{\extrarowheight}{1pt} 

\caption{Ablation study on dSprites regression tasks comparing different uncertainty quantification methods. Results are presented as the sum of MAE across three regression targets. \xmark indicates no uncertainty used (MSE loss for training), GP denotes Gaussian Processes, and DER represents Deep Evidential Regression.}
\begin{tabular}{l|c|cccccc|c}
\toprule
Method & Uncertainty & C $\rightarrow$ N & C $\rightarrow$ S & N $\rightarrow$ C & N $\rightarrow$ S & S $\rightarrow$ C & S $\rightarrow$ N  & Avg \\
\midrule
MMD \cite{long2015learning} & \xmark & 0.70 & 0.77 & 0.12 & 0.50 & 0.06 & 0.11 &0.38 \\
\midrule
\textbf{UGA-Post. MMD}& GP & 0.05 & 0.21 & 0.07 & 0.55 & 0.03 & 0.04 & 0.16 \\
\textbf{UGA-Post. MMD} & DER & 0.04 & 0.10 & 0.04 & 0.16 & 0.03 &  0.03 & 0.07\\
\midrule
\textbf{UGA-Feat. MMD} & GP & 0.05 & 0.19 & 0.05 & 0.40 & 0.03 & 0.03 & 0.13 \\
\textbf{UGA-Feat. MMD} & DER & \textbf{0.04} & \textbf{0.10} & \textbf{0.02} & \textbf{0.16} & \textbf{0.02} & \textbf{0.02} & \textbf{0.06} \\
\bottomrule
\end{tabular}

\label{tab:gp_vs_der}
\end{table}

The results in Table \ref{tab:gp_vs_der} clearly demonstrate the advantages of incorporating uncertainty in the domain adaptation process. Both Gaussian Process (GP) and Deep Evidential Regression (DER) uncertainty frameworks significantly improve upon the standard MMD approach across all transfer tasks. However, DER consistently outperforms GP, particularly in challenging transfer scenarios such as N → S and S → N.
This superior performance of DER can be attributed to its ability to model both aleatoric and epistemic uncertainty, providing a more comprehensive and robust guide for alignment for domain adaptation tasks. Furthermore, the results consistently show that feature alignment (UGA-Feat.) outperforms posterior alignment (UGA-Post.), regardless of the uncertainty quantification method used.

\section{Conclusion}

This paper introduces Uncertainty-Guided Alignement (UGA), a novel approach for Unsupervised Domain Adaptation in Regression tasks. By integrating predictive uncertainty into the feature alignment process, UGA effectively addresses the challenges arising from the interdependencies among regression features. We provide a theoretical foundation for expanding the embedding space to incorporate input information through uncertainty estimation, demonstrating how this approach overcomes the limitations of traditional feature alignment methods in regression contexts. Through comprehensive experiments on multiple computer vision benchmarks and a real-world battery life prediction task, we show that UGA consistently outperforms state-of-the-art methods across 52 diverse transfer tasks. Incorporating uncertainty into the alignment process significantly improves adaptation performance, particularly in challenging scenarios with substantial domain gaps, such as battery life prediction under extreme temperature variations. Moreover, UGA's well-calibrated uncertainty estimates offer valuable insights into the severity of domain shifts, facilitating more informed decision-making in critical applications such as prognostics and health management. We also demonstrate that while uncertainty estimation alone can effectively quantify the magnitude of domain shift,  our proposed alignment method further improves adaptation performance.

Although  UGA assumes similar target label ranges between domains, potentially limiting its applicability in certain scenarios, future research could explore leveraging uncertainty for test-time adaptation in regression tasks.

\bibliographystyle{IEEEtran}
\bibliography{cas-refs}

\end{document}